%% file: main.tex
\documentclass[sigconf]{acmart}

\def\BibTeX{{\rm B\kern-.05em{\sc i\kern-.025em b}\kern-.08emT\kern-.1667em\lower.7ex\hbox{E}\kern-.125emX}}

\copyrightyear{2019}
\acmYear{2019}
\setcopyright{acmcopyright}
\acmConference[IoTDI '19]{International Conference on Internet-of-Things Design and Implementation}{April 15--18, 2019}{Montreal, QC, Canada}
\acmBooktitle{International Conference on Internet-of-Things Design and Implementation (IoTDI '19), April 15--18, 2019, Montreal, QC, Canada}
\acmPrice{15.00}
\acmDOI{10.1145/3302505.3310070}
\acmISBN{978-1-4503-6283-2/19/04}

\usepackage{booktabs} 
\usepackage{amssymb}
\usepackage{subfigure}
\usepackage{multirow}

\renewcommand{\vec}[1]{\mathbf{#1}} 
\newcommand{\mat}[1]{\mathbf{#1}} 
\usepackage{color}

\newcommand{\blue}{}

\newcommand{\sect}{\textsection}

\DeclareMathOperator*{\argmin}{argmin}

\newtheorem{property}{Property}

\begin{document}

\title[On Lightweight Privacy-Preserving Collaborative Learning for IoT Objects]{On Lightweight Privacy-Preserving Collaborative Learning for Internet-of-Things Objects}

\author{Linshan Jiang}
\affiliation{
  \department{School of Computer Science and Engineering}
  \institution{Nanyang Technological University}
}
\email{linshan001@e.ntu.edu.sg}

\author{Rui Tan}
\affiliation{
  \department{School of Computer Science and Engineering}
  \institution{Nanyang Technological University}
}
\email{tanrui@ntu.edu.sg}

\author{Xin Lou}
\affiliation{
  \department{Advanced Digital Sciences Center}
  \institution{Illinois at Singapore Pte Ltd}
}
\email{lou.xin@adsc-create.edu.sg}

\author{Guosheng Lin}
\affiliation{
  \department{School of Computer Science and Engineering}
  \institution{Nanyang Technological University}
}
\email{gslin@ntu.edu.sg}

\input{abstract}

\ccsdesc[500]{Computer systems organization~Sensor networks}
\ccsdesc[500]{Computing methodologies~Supervised learning}
\ccsdesc[300]{Security and privacy~Domain-specific security and privacy architectures}

\keywords{Internet of Things, collaborative learning, privacy}

\maketitle

\input{intro}
\input{preliminaries}
\input{related}
\input{motivation}
\input{simulation}
\input{implementation}
\input{conclude}

\begin{acks}
  This research was funded by a Start-up Grant at Nanyang Technological University. We acknowledge the support of NVIDIA Corporation with the donation of two GPUs used in this research. We also acknowledge Dr. Yi Li for constructive discussions and Zhenyu Yan for managing the computation resources used in this paper.
\end{acks}

\bibliographystyle{ACM-Reference-Format}
\bibliography{main}

\end{document}

%% file: abstract.tex
\begin{abstract}
  The Internet of Things (IoT) will be a main data generation infrastructure for achieving better system intelligence. This paper considers the design and implementation of a practical privacy-preserving collaborative learning scheme, {\blue in which a curious learning coordinator trains a better machine learning model based on the data samples contributed by a number of IoT objects, while the confidentiality of the raw forms of the training data is protected against the coordinator.} Existing distributed machine learning and data encryption approaches incur significant computation and communication overhead, rendering them ill-suited for resource-constrained IoT objects. We study an approach that applies independent Gaussian random projection at each IoT object to obfuscate data and trains a deep neural network at {\blue the coordinator} based on the projected data from the IoT objects. This approach introduces light computation overhead to the IoT objects and moves most workload to {\blue the coordinator that can have sufficient computing resources. Although the independent projections performed by the IoT objects address the potential collusion between the curious coordinator and some compromised IoT objects, they significantly increase the complexity of the projected data. In this paper, we leverage the superior learning capability of deep learning in capturing sophisticated patterns to maintain good learning performance.} Extensive comparative evaluation shows that this approach outperforms other lightweight approaches that apply additive noisification for differential privacy and/or support vector machines for learning in the applications with light data pattern complexities.
\end{abstract}


%% file: intro.tex
\section{Introduction}
\label{sec:intro}

The recent research advances of machine learning have led to performance breakthroughs of various tasks such as image classification, speech recognition, and language understanding. The drastically increasing amount of data generated by the Internet of Things (IoT) will further foster machine learning performance and enable new applications in various domains.
In particular, the {\em collaborative learning}, which builds a machine learning model (e.g., a supervised classifier) based on the training data contributed by many {\em participants}, is a desirable and empowering paradigm for smarter IoT systems. By leveraging on the much increased volume of training data and coverage of data patterns, collaborative learning will approach the intelligence of a crowd and improve the learning performance beyond that achieved by any single participant alone.
Moreover, a resource-rich learning {\em coordinator} (e.g., a desktop-class edge device or a cloud computing service) allows the execution of advanced, compute-intensive machine learning algorithms to capture deeper structures in the aggregated data, whereas the participants (e.g., IoT objects) are often resource-constrained and incompetent for intensive computation. By contributing training data, the individual participants will enjoy the improved machine intelligence in return.

However, the data contributed by the participants may contain privacy-sensitive information. On Internet, various online services (e.g., webmail and social networking) generally collect and analyze the user data in the raw forms.
In this scheme, the users risk privacy leak due to potential inadvertent actions by the service providers and/or targeted cyber-attacks from the external. This risk has been evidenced by several recent large-scale user privacy leak incidents \cite{FB2018,Equifax,Lindsey2018}.
Data anonymization can mitigate the concern; but it is inadequate for privacy preservation, because cross correlations among different databases may be used to re-identify data \cite{narayanan2006break}. Moreover, the correlations between different properties of anonymous individuals (e.g., race, income, political views, etc.) can be exploited to identify an interested user group to target for advertisement and advocacy. In the coming era of IoT with many smart objects penetrating into our private space and time, the current raw data collection approach will only raise large privacy concerns and may potentially violate relevant laws such as the recent General Data Protection Regulation in European Union. Therefore, to be successful, the IoT-driven collaborative learning applications must be privacy-preserving.




Privacy-preserving collaborative learning (PPCL) has received increasing research recently under the enterprise settings, where the participants are entities with rich computing resources. The existing approaches can be broadly classified into two categories. The first category of approaches \cite{Hamm15,Shokri15,mcmahan2016communication,Phong17,Bonawitz17} follows the distributed machine learning (DML) scheme, such that the participants do not need to transmit the training data to the coordinator. {\blue The recently proposed {\em federated learning} \cite{mcmahan2016communication} is a type of DML.} In the second category of approaches \cite{graepel2012ml,gilad2016cryptonets,chabanne2017privacy}, each participant applies the homomorphic encryption on the data before being transmitted to the coordinator such that the training and inference can be performed on ciphertexts. However, for resource-constrained IoT objects, these DML and data encryption approaches incur significant and even prohibited compute overhead.
The DML will require the participants to execute machine learning algorithms to train local models, which is often too compute-intensive for IoT objects.
Moreover, the iterative communication rounds of DML introduce large communication overhead. Currently, the homomorphic encryption algorithms are still too compute-intensive to be realistic for resource-constrained devices (cf.~\sect\ref{sec:implementation}). Therefore, these existing approaches are ill-suited or unpractical for the resource-constrained smart objects beneath the IoT edge.

In this paper, we study the design and implementation of a PPCL approach that is lightweight for resource-constrained participants, while keeping privacy-preserving against a honest-but-curious learning coordinator. The coordinator can be a cloud server or a resource-rich edge device, e.g., access points, base stations, network routers, etc.
We propose to apply (1) multiplicative {\em Gaussian random projection} (GRP) at the resource-constrained IoT objects to obfuscate the contributed training data and (2) {\em deep learning} at the coordinator to address the much increased complexity of the data patterns due to the GRP. Specifically, each participant uses a private, time-invariant but randomly generated Gaussian matrix to project each plaintext training data vector and transmits the result to the coordinator. GRP gives several privacy preservation properties of (1) the computational difficulty for the coordinator to reconstruct the plaintext without knowing the Gaussian matrix \cite{liu2006random,rachlin2008secrecy}, and (2) quantifiable plaintext reconstruction error bounds even if the coordinator obtains the Gaussian matrix \cite{liu2006random}. From a system perspective, GRP is computationally lightweight and does not increase the data volume. Thus, GRP is a practical privacy protection method suitable for resource-constrained IoT objects. Regarding GRP's impact on the design of the machine learning algorithms, the random projection can be viewed as a process of mapping the original data vectors to
some domain in which the data vectors in different classes are less separatable. If the original data vectors are readily separatable (that is, they are features), the inverse of the Gaussian matrix can be considered as a linear feature extraction matrix. With the deep learning's unsupervised feature learning capability, this inverse matrix can be implicitly captured by the trained deep model. Thus, we conjecture that the randomly projected training samples can still be used by the coordinator to build the deep model for classification.

To achieve robustness of the privacy preservation against the collusion between any single participant and the curious learning coordinator, each participant should generate its own Gaussian matrix independently. However, this presents a {\blue main} challenge on the PPCL system's scalability with respect to the number of participants (denoted by $N$). Specifically, assuming that the training data samples for each class are horizontally distributed among the participants, the number of data patterns for a class will increase from one in the plaintext domain to $N$ in the projection data domain. This increased pattern complexity is to be addressed by the strong learning capability of deep learning. Thus, in the proposed PPCL approach, most of the computational workload is offloaded to the resourceful coordinator at the edge or in the cloud, unlike the existing DML and homomorphic encryption approaches that introduce significant or prohibitive compute overhead to the smart objects beneath the IoT edge.

To understand the effectiveness of the GRP approach and its scalability with the number of participants and the pattern complexity of the training data, we conduct extensive evaluation to compare GRP with several other lightweight PPCL approaches. The evaluation is based on two example applications with low and moderate pattern complexities, i.e., handwritten digit recognition and spam e-mail detection. The baseline approaches include various combinations between (1) multiplicative GRP versus additive noisification for differential privacy (DP) at the participants, and (2) deep neural networks (DNNs), {\blue including the multilayer perceptron (MLP) and convolutional neural network (CNN),}
versus support vector machines (SVMs) at the coordinator. The results show that, for the two example applications,
the proposed GRP-DNN approach can support up to hundreds of participants without sacrificing the learning performance much, whereas the GRP-SVM approach may fail to capture the projected data patterns and the performance of the DP-DNN approach is susceptible to additive noisification. The results of this paper suggest that GRP-DNN is a practical PPCL approach for resource-constrained IoT objects observing data with low- or moderate-complexity patterns.


{\blue We also implement GRP-DNN, Crowd-ML \cite{Hamm15} (a federated learning approach based on shallow learning), and CryptoNets \cite{gilad2016cryptonets} (a homomorphic encryption approach) on a testbed of 14 IoT devices. Experiments show that, compared with GRP-DNN, Crowd-ML incurs 350x compute overhead and 3.5x communication overhead to each IoT device. Deep federated learning will only incur more compute overhead. CryptoNets incurs 2.6 million times higher compute overhead to the IoT device, compared with GRP.}

The remainder of this paper is organized as follows. \sect\ref{sec:background} introduces the background and preliminaries. \sect\ref{sec:related} reviews related work. \sect\ref{sec:mov} states the problem and overviews our approach. \sect\ref{sec:evaluation} presents the learning performance evaluation for various lightweight PPCL approaches. \sect\ref{sec:implementation} presents the benchmark results of GRP-DNN, Crowd-ML, and CryptoNets on the testbed.
\sect\ref{sec:conclude} concludes this paper.


%% file: preliminaries.tex
\section{Background and Preliminaries}
\label{sec:background}



\subsection{Supervised Collaborative Learning}
\label{subsec:background}

Supervised machine learning has two phases, i.e., the learning phase and the classification phase.
We now formally describe the collaborative learning scheme. The trained classifier, denoted by $h(\vec{x} | \boldsymbol\theta )$, can classify a $d$-dimensional data vector $\vec{x} \in \mathbb{R}^d$ to be one of a finite number of classes represented by a set $\mathcal{C}$. The learning process determines the parameter $\boldsymbol\theta$ based on the training data. Let $N$ denote the number of participants of the collaborative learning. Let $\mathcal{D}_i$ denote a set of $M_i$ training data samples generated by the participant $i$, i.e., $\mathcal{D}_i = \{(\vec{x}_{i,j}, y_{i,j}) | j \in [1, M_i], y_{i,j} \in \mathcal{C}\}$, where $\vec{x}_{i,j}$ is the training data vector and $y_{i,j}$ is the corresponding class label. For a training data sample $(\vec{x}, y)$, denote by $l(h(\vec{x} | \boldsymbol\theta), y)$ the loss function. The collaborative learning solves the following problem to determine the optimal classifier parameter denoted by $\boldsymbol\theta^*$:
\begin{equation}
\boldsymbol\theta^* = \argmin_{\boldsymbol\theta} \sum_{i = 1}^{N} \frac{1}{M_i} \sum_{j=1}^{M_i} l\left( h \left( \vec{x}_{i,j} | \boldsymbol\theta \right), y_{i,j} \right) + \lambda \| \boldsymbol\theta \|^2,
\label{eq:learning}
\end{equation}
where the $\lambda \| \boldsymbol\theta \|^2$ is a regularization term. With $\boldsymbol\theta^*$, the classification for a test data sample $\vec{x}$ is to compute $h(\vec{x} | \boldsymbol\theta^*)$.

A simple approach is to collect all the plaintext training data to the coordinator and solve Eq.~(\ref{eq:learning}). However, this approach raises the concern of privacy breach, as the raw training data are generally privacy-sensitive. The problem of solving Eq.~(\ref{eq:learning}) without threatening the participants' privacy contained in $\mathcal{D}_i$, $i = 1, \ldots, N$, is called PPCL. Existing approaches to PPCL will be reviewed in \sect\ref{sec:related}.


\subsection{Random Gaussian Projection (GRP)}
\label{subsec:random-projection}

This section reviews two properties of GRP.
Let $\mathbf{R} \in \mathbb{R}^{k \times d}$ represent a random Gaussian matrix, i.e., each element in $\mathbf{R}$ is drawn independently from the normal distribution $\mathcal{N}(0, \sigma^2)$. GRP has the following two properties \cite{liu2006random}:

\begin{property}
  For data vectors $\vec{x}_1$, $\vec{x}_2$ and their projections $\vec{y}_1 = \frac{1}{\sqrt{k} \sigma}\mathbf{R}\vec{x}_1$, $\vec{y}_2 = \frac{1}{\sqrt{k} \sigma} \mathbf{R} \vec{x}_2$, the dot product and Euclidean distance between $\vec{y}_1$ and $\vec{y}_2$ are unbiased estimates of those between $\vec{x}_1$ and $\vec{x}_2$, i.e., $\mathbb{E} \left[ \vec{y}_1^\intercal \vec{y}_2 \right] = \vec{x}_1^\intercal \vec{x}_2$ and $\mathbb{E} \left[ \| \vec{y}_1 - \vec{y}_2 \|_2^2 \right] = \| \vec{x}_1 - \vec{x}_2 \|_2^2$. The estimation error bounds are $\mathrm{Var}[\vec{y}_1^\intercal \vec{y}_2] \le \frac{2}{k}$ and $\mathrm{Var}\left[ \|\vec{y}_1 - \vec{y}_2\|_2^2 \right] \le \frac{32}{k}$.
  \label{property:1}
\end{property}

\begin{property}
  Given a Gaussian matrix instance $\mathbf{R} \in \mathbb{R}^{k \times d}$ where $k < d$ and the projection $\vec{y} = \frac{1}{\sqrt{k}\sigma} \mathbf{R} \vec{x}$, the minimum norm estimate of $\vec{x}$, denoted by $\hat{\vec{x}}$, is an unbiased estimate of $\vec{x}$, i.e., $\mathbb{E}\left[ \hat{\vec{x}} \right] = \vec{x}$. The estimation error for the $i$th element of $\vec{x}$ is $\mathrm{Var}[x_i] = \frac{2}{k}x_i^2 + \frac{1}{k} \sum_{j, j \neq i} x_j^2$.
  \label{property:2}
\end{property}

{\blue Based on Property~\ref{property:1}, the study \cite{liu2006random} shows that a trained SVM classifier can be transferred to classify the projected data. In a recent study \cite{wojcik2018training}, a random projection layer that can be implemented by GRP is added to an MLP for dimension reduction. Such design is also based on Property~\ref{property:1}. However, the studies \cite{liu2006random,wojcik2018training} do not address collaborative learning and privacy.}

{\blue The estimation error given by Property~\ref{property:2} will be used in the later sections of this paper to measure the degree of privacy protection provided by our proposed approach.}

%% file: related.tex
\section{Related Work}
\label{sec:related}

Existing PPCL approaches can be classified into two categories, i.e., {\em distributed machine learning} (\sect\ref{subsubsec:distributed-learning}) and {\em training data encryption or obfuscation} (\sect\ref{subsubsec:data-encryption}). \sect\ref{subsubsec:other-related} reviews other related work.

\subsection{Distributed Machine Learning (DML)}
\label{subsubsec:distributed-learning}

DML approaches exploit the computing capability of the participants to solve Eq.~(\ref{eq:learning}) using some variant of stochastic gradient descent (SGD) in a distributed manner. During the learning process, the training data samples are not transmitted. The studies \cite{Hamm15,Shokri15,mcmahan2016communication,mcmahan2018learning} share the similar idea of exchanging gradients and classifier parameters among the participants, which is coordinated by the coordinator. Specifically, in the Crowd-ML approach \cite{Hamm15}, a participant checks out the global classifier parameters $\boldsymbol\theta$ from the coordinator and computes the gradients using its own training data. Then, the participants transmit the gradients to the coordinator that will update $\boldsymbol\theta$.
In \cite{Shokri15}, each participant trains a local deep model using SGD and uploads a selected portion of gradients to the coordinator for combining. Then, each participant downloads a selected portion of the global gradients to update its local deep model.
As the exchanged gradients and classifier parameters may still contain privacy, the approaches \cite{Hamm15,Shokri15} add random noises to the exchanged values for differential privacy \cite{Dwork06}.
In the {\em federated learning} scheme \cite{mcmahan2016communication}, the coordinator periodically pulls the deep models trained by the participants locally based on their training data and returns an average deep model to the participants. {\blue In \cite{mcmahan2018learning}, the participant adds random noises to the deep model parameters before being sent to the coordinator for privacy protection in the federated learning process.}

However, the above DML approaches have the following limitations. First, the local training introduces computation overhead to the participants. Training a DNN locally may be infeasible for resource-constrained IoT objects.
Second, DML approaches often require many iterations for the learning algorithm to converge, which may incur a high volume of data traffic between each participant and the coordinator. In \sect\ref{sec:implementation}, we will show this by comparing the Crowd-ML \cite{Hamm15} and our proposed approach. Third, as shown recently in \cite{Hitaj17}, generative adversarial networks can generate prototypical training data samples based on the exchanged gradients and model parameters, weakening the privacy preservation claimed in \cite{Shokri15,mcmahan2016communication}. In \cite{Phong17} and \cite{Bonawitz17}, homomorphic encryption and secure aggregation have been applied to enhance the privacy preservation of the approach in \cite{Shokri15} and the federated learning in \cite{mcmahan2016communication}, respectively. With these enhancements, only the encrypted gradients \cite{Phong17} and aggregate model update \cite{Bonawitz17} are revealed to the honest-but-curious coordinator. However, these privacy enhancements further increase the computation overhead of each participant, making it more unsuitable for resource-constrained IoT objects.





\subsection{Training Data Encryption/Obfuscation}
\label{subsubsec:data-encryption}

Different from the DML approaches that transmit classifier's parameters, the approaches \cite{graepel2012ml,liu2012cloud,shen2018privacy} transmit the encrypted or obfuscated training data to the coordinator to solve Eq.~(\ref{eq:learning}). {\blue The approach proposed in this paper also belongs to this category. In the following, we review each of \cite{graepel2012ml,liu2012cloud,shen2018privacy} and then discuss our new design to overcome their shortcomings.}

In \cite{graepel2012ml}, homomorphic encryption is integrated with a Linear Means classifier and Fisher's Linear Discriminant classifier. During both the training and classification phases, the participant transmits the homomorphically encrypted data vector to the coordinator. However, homomorphic encryption results in intensive computation and increased volume of data transmissions (cf.~\sect\ref{sec:implementation}). Thus, although the homomorphic encryption approach provides provable confidentiality protection, it is inefficient and unrealizable on many resource-constrained IoT platforms.

To reduce the computation and communication overheads, Liu et al. \cite{liu2012cloud} propose a data obfuscation approach based on random projection. Specifically, the participant $i$ independently generates a random matrix $\mathbf{R}_i$ and transmits the obfuscated training dataset $\{(\mathbf{R}_i\vec{x}_{i,j}, y_{i,j}) | j \in [1, M_i]\}$ to the coordinator.
However, different from Property~\ref{property:1} in \sect\ref{subsec:random-projection} that requires the same projection matrix, the approach \cite{liu2012cloud} uses distinct projection matrices for different participants and thus no longer preserves the Euclidean distance, i.e., $\| \mathbf{R}_{u} \vec{x}_{u,p} - \mathbf{R}_{v} \vec{x}_{v,q} \| \neq \| \vec{x}_{u,p} - \vec{x}_{v,q} \|$. This will result in poor training performance for distance-based classifiers, such as $k$-nearest neighbors and SVM. To address this issue, the study \cite{liu2012cloud} designs a regression phase before the learning phase. Specifically, the coordinator sends a number of {\em public data vectors} $\{\vec{z}_k | k=1, 2, \ldots \}$ to all participants and the participant $i$ returns the projected data $\{ \mathbf{R}_i \vec{z}_k | k=1, 2, \ldots \}$. Based on the original and projected public data vectors, a regress function $f_{uv}(\cdot, \cdot)$ for each participant pair $(u,v)$ is learned such that $f_{uv}(\mathbf{R}_u \vec{x}_{u,p}, \mathbf{R}_v \vec{x}_{v,q}) \simeq \| \vec{x}_{u,p} - \vec{x}_{v,q} \|$. As a result, the distance-based classifiers can be trained in the domain of obfuscated data by using the learned regress functions to compute distances during the training phase.

However, the approach \cite{liu2012cloud} has two shortcomings. First, it is only applicable to distance-based classifiers. These conventional classifiers do not scale well with the volume of the training data and the complexity of the data patterns \cite{suykens2003}. It is desirable to support the DNNs that give the state-of-the-art learning performance in a range of applications. Second, obfuscating the public data vectors and returning the results may incur known-plaintext attacks and engenders a clear privacy breaching concern. For instance, a proactively curious coordinator may use a public data vector $\vec{z}_k = [1, 0, 0, \ldots, 0]^\intercal$ to extract the first column of $\mathbf{R}_i$. Other columns of $\mathbf{R}_i$ can be similarly extracted by using specific public data vectors. Even without using these specific public data vectors, in general, the private random projection matrix $\mathbf{R}_i$ can be estimated using regression analysis based on a number of public data vectors and the corresponding projections.




The study \cite{shen2018privacy} also uses random projection to obfuscate the data vector $\vec{x}$ in training and executing a Sparse Representation Classifier. However, all participants use the same random projection matrix, rendering the system vulnerable to the collusion between any single participant and the coordinator.

{\blue Different from \cite{shen2018privacy}, each participant in our approach uses its own private random project matrix, rendering the collusion futile. Different from \cite{liu2012cloud}, our approach uses DNNs and leverages on the deep learning capability to avoid the regression phase that is vulnerable to the known-plaintext attacks. Different from \cite{graepel2012ml} that is too compute-intensive for IoT objects, our approach uses GRP that introduces light computation overhead only.}

\subsection{Other Related Work}
\label{subsubsec:other-related}

In CryptoNets \cite{gilad2016cryptonets}, the computation of each neuron in a neural network trained using plaintext data is performed in the domain of homomorphic encryption. During the classification phase, the participant sends the homomorphically encrypted data to the coordinator for classification. The work \cite{chabanne2017privacy} extends \cite{gilad2016cryptonets} to support more hidden layers. However, these studies \cite{gilad2016cryptonets,chabanne2017privacy} address {\em privacy-preserving classification outsourcing} (i.e., offloading the classification computation to a honest-but-curious entity), rather than the collaborative learning addressed in this paper. The training in \cite{gilad2016cryptonets,chabanne2017privacy} is performed based on plaintext data. Moreover, the homomorphic encryption is too computation intensive for resource-constrained IoT devices, which will be shown in \sect\ref{sec:implementation}.




The {\em differentially private machine learning} (DPML) \cite{Abadi16,chaudhuri2009privacy,song2013stochastic} builds a classifier that cannot be used to infer the training data. The training of the classifier is based on plaintext data. For DNNs, DPML can be achieved by perturbing the gradients in each iteration of the SGD with additive noises \cite{Abadi16,song2013stochastic}. DPML and PPCL address different problems, i.e., PPCL preserves the privacy of the training data against the honest-but-curious coordinator who builds the classifier, whereas DPML trusts the classifier builder and preserves the privacy of the training data against the curious user of the classifier. Thus, in DPML, the plaintext training dataset is available to the classifier builder; differently, in PPCL, only encrypted or obfuscated training data is made available to the classifier builder (i.e., the learning coordinator).







%% file: motivation.tex
\section{Problem Statement and Approach}
\label{sec:mov}

In this section, we state the PPCL problem in \sect\ref{subsec:problem-statement} and present the proposed GRP approach in \sect\ref{subsec:approach-overview}. \sect\ref{subsec:example} provides two illustrating examples for insights into understanding the effect of GRP on training DNN-based classifiers. \sect\ref{subsec:alternatives} discusses two other alternative approaches for lightweight PPCL and their limitations.

\subsection{Problem Statement}
\label{subsec:problem-statement}

This section states the problem addressed in this paper. \sect\ref{subsubsec:system-model} defines the system model; \sect\ref{subsubsec:threat-model} defines the threat and privacy models; \sect\ref{subsubsect:several-issues} discusses several relevant issues.

\subsubsection{System model}
\label{subsubsec:system-model}

\begin{figure}
  \centering
  \includegraphics{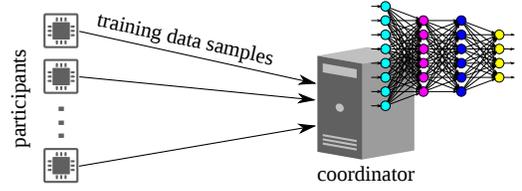}
  \caption{A collaborative learning system.}
  \label{fig:system}
\end{figure}

In this paper, we consider a PPCL system with $N$ resource-constrained {\em participants}  and an honest-but-curious {\em coordinator} with sufficient computation power. Fig.~\ref{fig:system} illustrates the system. During the learning phase, the participants contribute training data samples to build a supervised classifier.
As discussed in \sect\ref{subsec:background}, the training dataset $\mathcal{D}_i$ contributed by the participant $i$ consists of $M_i$ data vectors $\{\vec{x}_{i,j} | j \in [1, M_i]\}$ and the corresponding class labels $\{y_{i,j} | j \in [1, M_i] \}$.
As the learning process is often compute-intensive, most of the learning computation should be accomplished by the coordinator. In this paper, we focus on addressing the problem of building an effective supervised classifier while protecting certain privacy contained in the data vectors.




\subsubsection{Threat and privacy models}
\label{subsubsec:threat-model}

The privacy concern regarding the data vectors is primarily due to that the data vectors may contain information beyond the classification objective in question. For example, consider a PPCL system for training a classifier to recognize human body activity (e.g., sitting, walking, climbing stairs, etc). The recognition is based on various body signals (e.g., motion, heart rate, breath rate, etc) that are captured by wearable sensors. However, the raw body signals can also be used to infer health statuses of the participants and even pinpoint the patients of certain diseases. In this paper, we adopt the following threat and privacy models.




{\blue

  \vspace{0.5em}
  {\em Threat model:} It consists of the following two aspects:}
\begin{itemize}
\item {\em Honest-but-curious coordinator:} We assume that the coordinator will honestly coordinate the collaborative learning process, aiming to train the best supervised classifier. Thus, it will neither tamper with any data {\blue collected from or transmitted to} the participants.
  However, the coordinator is curious about the participants' privacy contained in the training data vectors.
  \item {\em Potential collusion between participants and coordinator:} We assume that the participants are not trustworthy in that they may collude with the coordinator in finding out other participants' privacy contained in the data vectors. The colluding participants are also honest, i.e., they will faithfully contribute their training data to improve the supervised classifier. The design of the PPCL system should keep the privacy preservation for a participant when any or all other participants are colluding with the coordinator.
  \end{itemize}

  {\blue
    \vspace{0.5em}
    {\em Privacy model:} The raw form of each data vector is the participant's privacy to be protected. The error in estimating the data raw form by the coordinator can be used as a metric to measure the degree of privacy protection. Data form confidentiality is an immediate and basic privacy requirement in many applications.
    \vspace{0.5em}
}




\subsubsection{Several other issues}
\label{subsubsect:several-issues}

In the following, we discuss three issues that are related to privacy protection:
\begin{itemize}
\item {\em Training data anonymization:} We aim to support anonymization of the training data. That is, the coordinator should not expect to know the participant's identity for any received training data sample. Moreover, the coordinator cannot determine whether any two training data samples are from the same participant. To achieve the above strong anonymity, the training data samples can be transmitted in separate sessions via an anonymous communication network \cite{danezis2008survey}. {\blue Moreover, the transmissions of the data samples from all participants can be interleaved randomly, such that the coordinator cannot associate the data samples from the same participant by their arrival times.} Note that the training data anonymization requirement is not mandatory, because the anonymous communication may incur large overhead for some resource-constrained IoT objects. However, the design of our PPCL approach will not leverage the participants' identities to support data anonymization.
\item {\em Label privacy:} The class labels $\{y_{i,j} | j \in [1, M_i] \}$ may also contain information about the participant. In this paper,  we do not consider label privacy because the participant willingly contributes the labeled data vectors and should have no expectation of privacy regarding labels. In practice, several means can be taken to mitigate the concern of label privacy leak. First, the training data anonymization mitigates the concern during the learning phase.
Second, during the classification phase, if the participant has sufficient processing capability to perform the classification computation, the coordinator may send the trained model to the participant for local execution. Existing studies have enabled the execution of deep models on personal and low-end devices \cite{yao2017deepiot,huynh2017deepmon}. Low-power inference chips (e.g., Google's Edge TPU \cite{edge-tpu}) will further enhance low-end devices' capabilities in executing classification models. Note that the studies \cite{yao2017deepiot,huynh2017deepmon} and the inference chips are not to support the much more compute-intensive training.
\item {\blue {\em Other privacy models:} Differential privacy \cite{Dwork06} aim to achieve indistinguishability of different data vectors is another widely used quantifiable privacy definition. However, as discussed in \sect\ref{subsec:alternatives} and evaluated in \sect\ref{sec:evaluation}, the additive noisification implementation of differential privacy is ill-suited for PPCL.}
\end{itemize}

\subsection{Gaussian Random Projection Approach}
\label{subsec:approach-overview}

Existing DML and homomorphic encryption approaches incur significant computation and communication overhead due to the many computation/communication rounds and data volume swell. In \sect\ref{sec:implementation}, we will provide benchmark results to show this. Thus, these approaches are not promising for resource-constrained participants.
This section describes a GRP-based approach that is computationally lightweight and communication efficient for the participants. The overview of our approach is presented as follows.

At the system initialization, each participant $i$ independently generates a random Gaussian matrix $\mathbf{R}_i \in \mathbb{R}^{k \times d}$, where $d$ is the dimension of the data vector.
During the learning phase, the participant $i$ keeps $\mathbf{R}_i$ secret and uses it to project all the training data vectors. The participant $i$ transmits the projected training dataset $\mathcal{D}_i = \{\mathbf{R}_i\vec{x}_{i,j}, y_{i,j} | j \in [1, M_i], y_{i,j} \in \mathcal{C} \}$ to the coordinator. After collecting all projected training datasets $\mathcal{D}_i$, $i = 1, \ldots, N$, the coordinator applies deep learning algorithms to train the classifier $h(\cdot | \boldsymbol\theta^*)$. During the classification phase, the participant $i$ still uses $\mathbf{R}_i$ to project the test data vector $\vec{x}$ and obtains the classification result $h(\mathbf{R}_i\vec{x} | \boldsymbol\theta^*)$. As discussed in \sect\ref{subsec:problem-statement}, the classification computation can be carried out at the participant or the coordinator, depending on whether the participant is capable of executing the trained deep model. In our approach, each participant independently generates its random projection matrix to counteract the collusion between participants and coordinator. Now, we explain the two key components of our approach: GRP and deep learning on projected data.

\subsubsection{Gaussian random projection}
\label{subsubsec:projection}


In this work, we adopt Gaussian matrices. Specifically, each element of $\mathbf{R}_i$ is sampled independently from the standard normal distribution \cite{Ailon09}. The rationale of choosing Gaussian matrices will be explained in \sect\ref{subsubsec:10d-example}.
We set the row dimension of $\mathbf{R}_i$ smaller than or equal to its column dimension, i.e., $k \le d$. Thus, the GRP can also compress the data vector. We define the compression ratio as $\rho = d / k$. The understanding regarding the admission of compression into the training data projection is as follows. From the compressive sensing theory \cite{candes2008introduction}, a sparse signal can be represented by a small number of linear projections of the original signal and recovered faithfully. Therefore, in the compressively projected data vector, the feature information still exists, provided that the adopted compression ratio is within an analytic bound \cite{candes2008introduction}.
In \sect\ref{sec:evaluation}, we will evaluate the impact of the compression ratio $\rho$ on the learning performance.


\begin{figure*}
  \centering
  \subfigure[Original data]
  {
    \includegraphics{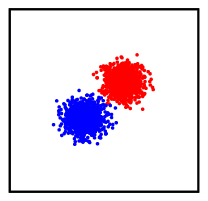}
   \label{fig:toy1-original}
  }
  \subfigure[Participant 1]
  {
    \includegraphics{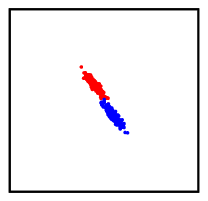}
    \label{fig:toy1-p1}
 }
  \subfigure[Participant 2]
 {
    \includegraphics{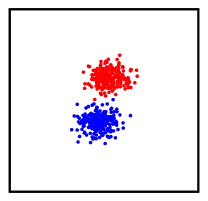}
    \label{fig:toy1-p2}
  }
  \subfigure[Participant 3]
  {
    \includegraphics{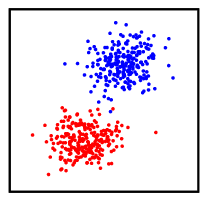}
   \label{fig:toy1-p3}
 }
  \subfigure[Participant 4]
  {
    \includegraphics{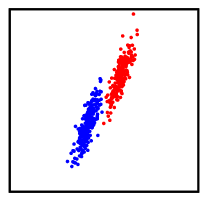}
    \label{fig:toy1-p4}
  }
  \subfigure[Coordinator]
  {
    \includegraphics{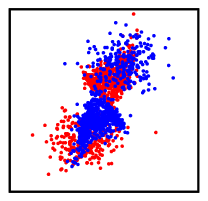}
    \label{fig:toy1-projection1}
  }
  \subfigure[Coordinator]
  {
    \includegraphics{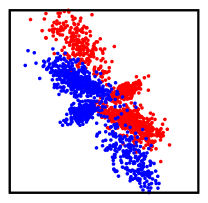}
    \label{fig:toy1-projection2}
  }
  \subfigure[Coordinator]
  {
    \includegraphics{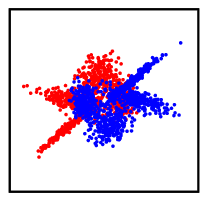}
    \label{fig:toy1-projection3}
  }
  \caption{Two-dimensional example. Original data vectors and projected data vectors (red: class 0; blue: class 1). The ranges for the $x$ and $y$ axes are $[-10,10]$.}
  \label{fig:random_projection}
\end{figure*}

With GRP, if $\mathbf{R}_i$ is kept confidential to the coordinator, it is computationally difficult (practically impossible) for the coordinator to generate a meaningful reconstruction of the original data vector from the projected data vector \cite{liu2006random,rachlin2008secrecy}. Thus, GRP protects the form of the original data.
In the worst case where the coordinator obtains $\mathbf{R}_i$, the estimation error given by Property~\ref{property:2} in \sect\ref{subsec:random-projection} can be used as a measure of privacy protection.
Random projection has been used as a lightweight approach to protect data form confidentiality in various contexts \cite{li2013compressed,tan2017joint,wang2013privacy,xue2017kryptein}.

\subsubsection{Deep learning on projected data}
\label{subsubsec:dl}

Feature extraction is a critical step of supervised learning. With the traditional {\em shallow learning}, the classification system designer needs to handcraft the feature. The emerging deep learning method \cite{lecun2015deep} automates the design of feature extraction by {\em unsupervised feature learning}, which is often based on a neural network consisting of a large number of parameters. Thus, the deep model is often a tandem of the feature extraction stage and the classification stage. For example, a convolutional neural network (CNN) for image classification consists of convolutional layers and dense layers, which are often considered performing the feature extraction and classification, respectively.

Our approach leverages on the unsupervised feature learning capability of deep learning to address the data distortion introduced by the GRP. We now illustrate this using a simple example system, in which there is only one participant and the projection matrix $\mathbf{R}$ is a square invertible matrix. Moreover, we make the following two assumptions to simplify our discussion. First, we assume that a linear transform $\boldsymbol\Psi \in \mathbb{R}^{f \times d}$ gives effective features of the data vectors, where $f$ is the feature dimension. That is, $\vec{f} = \boldsymbol\Psi \vec{x}$ is an effective representation of the data vector $\vec{x}$ for classification. Second, we assume that $\boldsymbol\Psi$ can be learned in the form of a neural network by the unsupervised feature learning. Now, we discuss the impact of the random projection on the unsupervised feature learning. After the projection, the data vector becomes $\mathbf{R} \vec{x}$.
Moreover, the linear transform $\boldsymbol\Psi \mathbf{R}^{-1}$ will be an effective feature extraction method, since $\vec{f} = \left( \boldsymbol\Psi \mathbf{R}^{-1} \right) \left( \mathbf{R} \vec{x} \right)$. It is reasonable to expect that the unsupervised feature learning can also build a neural network to capture the linear transform $\boldsymbol\Psi \mathbf{R}^{-1}$, similar to the unsupervised feature learning to capture the $\boldsymbol\Psi$ based on the plaintext training data $\vec{x}$. As a result, the deep model trained using the projected data can still classify future projected data vectors. In \sect\ref{subsec:example}, we will use a numerical example to illustrate this.

The above discussion based on linear features provides a basis for us to understand how the unsupervised feature learning helps address the distortion caused by the GRP. In practice, effective feature extractions are generally non-linear mappings. Neural network-based deep learning has shown strong capability in capturing sophisticated features beyond the above ideal linear features. In this paper, based on multiple datasets, we will investigate the effectiveness of deep learning to address the distortion caused by the GRP.



As discussed earlier, each participant independently generates a Gaussian matrix to counteract the potential collusion between participants and the coordinator. However, this introduces a challenge to deep learning, because the pattern for a class of projected data vectors from $N$ participants will be a composite of $N$ different patterns. Thus, intuitively, a deeper neural network and a larger volume of training data will be needed to well capture the data patterns with increased complexity due to the participants' independence in generating their projection matrices. We note that, the participants' independence also engenders the following possible situation that undermines the learning performance and leads to classification errors: $\mathbf{R}_u \vec{x}_u = \mathbf{R}_v \vec{x}_v$, where $\vec{x}_u$ and $\vec{x}_v$ are generated by participants $u$ and $v$ and belong to different classes. However, for high-dimensional data vectors, the probability of the above situation is low. The more complex data patterns due to the independent projection matrix generation will be the major challenge. In this paper, we conduct extensive experiments to assess how well deep learning can scale with the number of participants, compared with the traditional learning approaches.




\subsection{Illustrating Examples}
\label{subsec:example}

We use two examples to illustrate the intuitions discussed in \sect\ref{subsec:approach-overview}.

\subsubsection{A 2-dimensional example}


We consider a PPCL system with four participants (i.e., $N=4$) to build a two-class classifier. The original data vectors in the two classes follow two 2-dimensional Gaussian distributions with means of $[-2,-2]^\intercal$ and $[2,2]^\intercal$, and the same covariance matrix of $[1, 0; 0, 1]$. Fig.~\ref{fig:toy1-original} shows the plaintext data vectors generated by the four participants. From the figure, the plaintext data vectors of the two classes can be easily separated using a simple hyperplane. Each participant independently generates a Gaussian random matrix. Figs.~\ref{fig:toy1-p1}-\ref{fig:toy1-p4} show the projected data vectors of each participant. We can see that the patterns of the projected data vectors are different across the participants. Fig.~\ref{fig:toy1-projection1} shows the mixed projected data vectors received from all participants. Compared with Fig.~\ref{fig:toy1-original}, the pattern of the mixed projected data from all participants is highly complex. Moreover, no simple hyperplane can well divide the two classes.

We also generate two other sets of the random projection matrices for all participants. Figs.~\ref{fig:toy1-projection2} and \ref{fig:toy1-projection3} show the mixes of all participants' projected data vectors with the two sets of random projection matrices, respectively. Similarly, the pattern of the mixed projected data from all participants is highly complex.

We construct a classifier based on an MLP with two hidden layers of 30 and 40 rectified linear units (ReLUs), respectively. The input layer admits a 2-dimensional data vector, whereas the output layer consists of two ReLUs. The final classification result is generated using a softmax function based on the output layer's ReLU values. Moreover, we construct an SVM classifier as a baseline approach. We use LIBSVM \cite{libSVM} to implement the classifier. The SVM classifier uses radial basis function (RBF) kernel with two configurable parameters $C$ and $\lambda$. During the training phase, we apply grid search to determine the optimal settings for $C$ and $\lambda$.

First, we use disjoint subsets of the original data shown in Fig.~\ref{fig:toy1-original} to train and test the MLP and SVM classifiers. Both classifiers can achieve 99\% test accuracy. This shows that the MLP and the SVM are properly designed for the 2-dimensional data vectors.

Then, we use disjoint subsets of the randomly projected data shown in Fig.~\ref{fig:toy1-projection1} to train and test the MLP and SVM classifiers. Moreover, we also increase the number of participants in the PPCL system. Fig.~\ref{fig:toydata1} shows the test accuracy versus the number of participants. We can see that the MLP classifier always outperforms the SVM classifier. Moreover, the test accuracy decreases with the number of participants. This is because, with more participants, the pattern of the projected data becomes more complex, introducing challenges to both MLP and SVM. The test accuracy difference between MLP and SVM increases from 2\% to 7\%, when the number of participants increases from 4 to 20. This result is also consistent with the understanding that deep learning is more effective in capturing complex patterns than traditional learning.


\subsubsection{A 10-dimensional example}
\label{subsubsec:10d-example}

Now, we use another example system to understand the effect of deep learning's unsupervised feature learning capability in addressing the data distortion caused by the random projection. This example is a PPCL system with only one participant (i.e., $N=1$).
The original data vectors in two classes follow two 10-dimensional Gaussian distributions, with the $[-2,-2,\ldots,-2]^\intercal$ and $[2,2,\ldots,2]^\intercal$ as the respective mean vectors, and the 10-dimensional identity matrix as their identical covariance matrix.

\begin{figure}
  \centering
  \begin{minipage}[t]{.225\textwidth}
    \centering
    \includegraphics{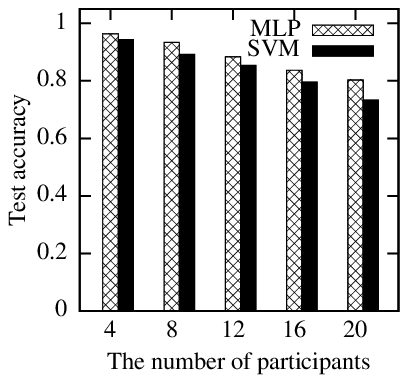}
    \caption{Test accuracy based on projected data vs. the number of participants.}
    \label{fig:toydata1}
  \end{minipage}
  \hspace{0.01\textwidth}
  \begin{minipage}[t]{.225\textwidth}
    \centering
    \includegraphics{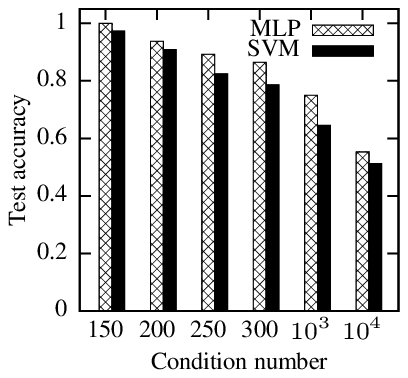}
    \caption{Test accuracy based on projected data vs. the condition number.}
    \label{fig:toydata2}
  \end{minipage}
\end{figure}

In our discussions in \sect\ref{subsubsec:dl}, we assume that the projection matrix $\mathbf{R}$ is invertible and the unsupervised feature learning tend to capture $\boldsymbol\Psi \mathbf{R}^{-1}$. As learning algorithms are based on numerical computation on the training data, an ill-conditioned matrix $\mathbf{R}$ will impede efficient fitting of $\boldsymbol\Psi \mathbf{R}^{-1}$. We verify this intuition by assessing the learning performance of the single-participant PPCL system using different $\mathbf{R}$ matrices with varying condition numbers.
Specifically, by following a method described in \cite{bierlaire1991iterative}, the participant generates a random square matrix $\mathbf{R}$ that has a certain condition number value.
The condition number is defined as $\|\mathbf{R}\|_F \| \mathbf{R}^+ \|_F$ \cite{paige1982lsqr}, where $\mathbf{R}^+$ denotes the pseudoinverse of $\mathbf{R}$ and $\|\cdot\|_F$ represents the Frobenius norm.
Fig.~\ref{fig:toydata2} shows the test accuracy of the MLP and SVM classifiers trained using data projected by $\mathbf{R}$ versus the condition number of $\mathbf{R}$. Note that a larger condition number means that the matrix is more ill-conditioned. We can see that the test accuracy decreases with the condition number, consistent with the intuition.

The study \cite{chen2005condition} analyzes the distribution of the condition numbers of Gaussian random matrices. The results show that a Gaussian random matrix is well-conditioned with a high probability. For instance, it is shown in \cite{chen2005condition} that for a $10 \times 5$ Gaussian random matrix, the probability that its condition number is larger than 100 is less than $6 \times 10^{-7}$. This is a basis for our choice of using Gaussian random matrices to project data.






\subsection{Alternative Approaches and Limitations}
\label{subsec:alternatives}

This section discusses two alternative approaches to PPCL and their limitations. These two alternatives will be used as the baseline approaches in our comparative performance evaluation in \sect\ref{sec:evaluation}.

\subsubsection{Non-collaborative learning}
\label{subsubsec:nc-learn}

If the data anonymity requirement {\blue is not enforced}, the coordinator can train a separate deep model based on the projected data vectors contributed by each participant. This alternative approach can address the challenge of the complex mixed patterns due to different random projection matrices adopted by different participants as illustrated in \sect\ref{subsec:example}. However, it loses the advantages of collaborative learning, i.e., the increased data volume and pattern coverage. From our evaluation in \sect\ref{sec:evaluation}, compared with our proposed approach, despite that this non-collaborative learning approach additionally uses the participant identity information, it yields inferior average accuracy.

\subsubsection{Differential privacy}
\label{subsubsec:dp}

Differential privacy (DP) \cite{Dwork06} is a rigorous information-theoretic approach to prevent leak of individual records by statistical queries on a database of these records. The $\epsilon$-DP \cite{Dwork06} is formally defined as follows: {\em A randomized algorithm $\mathcal{A}: \mathbb{D}\rightarrow \mathbb{R}^t$ gives $\epsilon$-DP if for all adjacent datasets $D_1 \in \mathbb{D}$ and $D_2 \in \mathbb{D}$ differing on at most one element, and all $S\subseteq Range(\mathcal{A})$, $\Pr(\mathcal{A}(D_1) \in S)\leq \exp(\epsilon)\cdot \Pr(\mathcal{A}(D_2) \in S)$.} The $\epsilon$, a positive real number, is a measure of privacy loss, i.e., a smaller $\epsilon$ implies better privacy. When $\epsilon$ is very small, $\Pr(\mathcal{A}(D_1) \in S) \simeq \Pr(\mathcal{A}(D_2) \in S)$ for all $S \subseteq Range(\mathcal{A})$, which means that the query results $\mathcal{A}(D_1)$ and $\mathcal{A}(D_2)$ are almost indistinguishable based on any ``test criterion'' of $S$.
The indistinguishability between the query results $\mathcal{A}(D_1)$ and $\mathcal{A}(D_2)$ decreases with $\epsilon$. The study \cite{Dwork061} develops an approach of adding Laplacian noises to implement $\epsilon$-DP. Specifically, for all function $\mathcal{F}: \mathcal{D} \rightarrow \mathbb{R}^t$, the randomized algorithm $\mathcal{A}(D) = \mathcal{F}(D) + [n_1, n_2, \ldots, n_t]^\intercal$ gives $\epsilon$-DP, where each $n_i$ is drawn independently from a Laplace distribution $\mathrm{Lap}(S(\mathcal{F})/\epsilon)$ and $S(\mathcal{F})$ denotes the global sensitivity of $\mathcal{F}$. Note that $\mathrm{Lap}(\lambda)$ denotes a zero-mean Laplace distribution with a probability density function of $f(x|\lambda)=\frac{1}{2\lambda}e^{\frac{|x|}{\lambda}}$; the global sensitivity is
\begin{equation*}
  S(\mathcal{F})=\max_{\forall D' \in \mathbb{D}, \forall D'' \in \mathbb{D}}||\mathcal{F}(D')-\mathcal{F}(D'')||_1.
\end{equation*}

Essentially, $\epsilon$-DP gives quantifiable indistinguishability of the query results based on different datasets. The $\epsilon$-DP framework has been applied in various privacy preservation problems in machine learning. As discussed in \sect\ref{subsubsec:distributed-learning}, the DML approaches to PPCL \cite{Hamm15,Shokri15} add random noises to the parameters exchanged between the participants and the coordinator to achieve $\epsilon$-DP. The original parameters can be viewed as deterministic query results of the training data. Adding random noises to the parameters ensures certain levels of indistinguishability between the noise-added parameters based on different training datasets.
The achieved $\epsilon$-DP mitigates the privacy concern that the curious coordinator may use the received parameters to infer the existence of particular data vectors in the training dataset.
However, these DML approaches \cite{Hamm15,Shokri15}
incur significant overhead to resource-constrained participants.
For PPCL based on resource-constrained participants, an approach to achieving $\epsilon$-DP is to add a Laplacian noise vector to the original data vector $\vec{x}$ and then transmit the noise-added data vector to the coordinator for building the classifier. By doing so, certain levels of indistinguishability between the noise-added data vectors based on different original data vectors are achieved.

Additive noisification and multiplicative GRP preserve different forms of privacy. Compared with protecting indistinguishability under the DP framework, we believe that protecting the confidentiality of the raw data form, which can be achieved by GRP, is a more immediate and basic privacy requirement in many applications.
The additive noisification, though achieving $\epsilon$-DP, falls short of protecting the confidentiality of the raw data form. Specifically, under the $\epsilon$-DP framework based on zero-mean Laplacian noises, a noise-added data vector can be considered an unbiased estimate of the original data vector with an estimation variance related to $\epsilon$. Thus, the coordinator always has a meaningful (i.e., unbiased) estimate of the raw data.
According to Property~\ref{property:2} in \sect\ref{subsec:random-projection}, this only happens to the GRP approach in the worst (and unrealistic) case that the projection matrix is revealed to the coordinator; other than the worst case, the coordinator cannot have a meaningful estimate of the raw data form. In the image classification case studies in \sect\ref{sec:evaluation}, we will show that when $\epsilon$ is small (i.e., good DP), the contents of the noise-added images can still be interpreted. In contrast, the projected images cannot be interpreted visually at all.


Applying $\epsilon$-DP to PPCL with resource-constrained participants also introduces the following two challenges.

\vspace{0.5em}
\noindent {\em Non-trivial computation overhead:} From the DP theory, an independent random noise vector should be generated and added to every data vector $\vec{x}$. However, random number generation is often a costly operation due to the use of various mathematical functions.
The continuous generation of Laplacian noises will incur non-trivial computation overhead for the resource-constrained participants. Differently, in our approach, the random projection matrix generation is a one-off overhead. The projection to compute $\mathbf{R}\vec{x}$ is a lightweight operation consisting of multiplications and additions only. {\blue Our previous work \cite{tan2017joint} has implemented the projection operation on an MSP430-based platform. Moreover, the projection can be sped up if a parallel computing chip (e.g., Google's Edge TPU \cite{edge-tpu}) is available.}



\vspace{0.5em}
\noindent {\em Learning performance degradation:} As discussed in \sect\ref{subsubsec:dl}, the projection matrix can be implicitly learned by the deep learning algorithms. Differently, the additive Laplacian noises to ensure $\epsilon$-DP can be considered neither a pattern nor an embedding that can be learned by learning algorithms. Thus, the Laplacian noises will only negatively affect the learning performance. Our evaluation in \sect\ref{sec:evaluation} shows that the Laplacian noises for achieving moderate $\epsilon$-DP significantly degrade the learning performance.

From the above discussions and the evaluation results in \sect\ref{sec:evaluation}, adding Laplacian noises to the training data for $\epsilon$-DP is not a promising approach to PPCL with resource-constrained participants.



%% file: simulation.tex
\section{Performance Evaluation}
\label{sec:evaluation}

{\blue In this section, we extensively compare the accuracy achieved by various approaches. The computation and communication overhead of these approaches will be profiled in \sect\ref{sec:implementation} based on their implementations on a testbed.}
  

\subsection{Evaluation Methodology and Datasets}

We conduct extensive evaluation to compare several approaches:
\begin{itemize}
\item {\bf GRP-DNN:} This is the proposed approach consisting of GRP at the participants and collaborative learning based on a DNN at the coordinator. The design or choice of the DNN model will be application specific. The DNN models and training algorithms are implemented based on PyTorch \cite{pytorch}.
\item {\bf GRP-SVM:} This baseline approach applies GRP at the participants and trains an SVM-based classifier at the coordinator. The SVM-based classifier is implemented using LIBSVM \cite{libSVM}. The classifier uses RBF kernel with two configurable parameters $C$ and $\lambda$. During the training phase, we apply grid search to determine the best settings for $C$ and $\lambda$. This grid search is often lengthy in time (e.g., several days).
\item {\bf GRP-NCL:} This is the non-collaborative learning (NCL) baseline approach described in \sect\ref{subsubsec:nc-learn}. It runs GRP at the participants and trains a separate DNN for each participant at the coordinator. Compared with other approaches, this approach additionally requires the identity of the participant for each training sample.
\item {\bf $\epsilon$-DP-DNN:} As described in \sect\ref{subsubsec:dp}, this approach implements $\epsilon$-DP by adding Laplacian noise vectors to the data vectors and performs collaborative deep learning based on a DNN at the coordinator.
\item {\bf $\epsilon$-DP-SVM:} This approach implements $\epsilon$-DP by adding Laplacian noise vectors to the data vectors and performs collaborative learning based on SVM at the coordinator.
\item {\blue {\bf CNN, SVM, MLP, ResNet-152:} These are the plain learning approaches based on the CNN, SVM, MLP, and ResNet-152 models, respectively. They do not protect any privacy.}
\end{itemize}

\begin{figure}
  \subfigure[Original images]
  {
    \includegraphics[width=.044\textwidth]{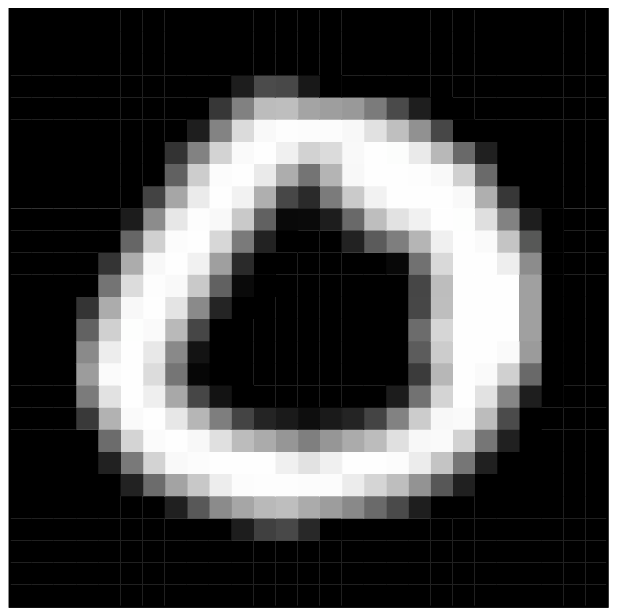}
    \includegraphics[width=.044\textwidth]{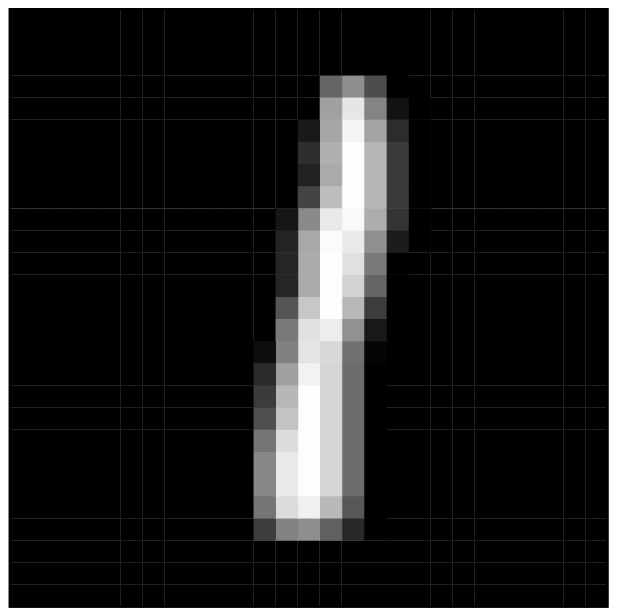}
    \includegraphics[width=.044\textwidth]{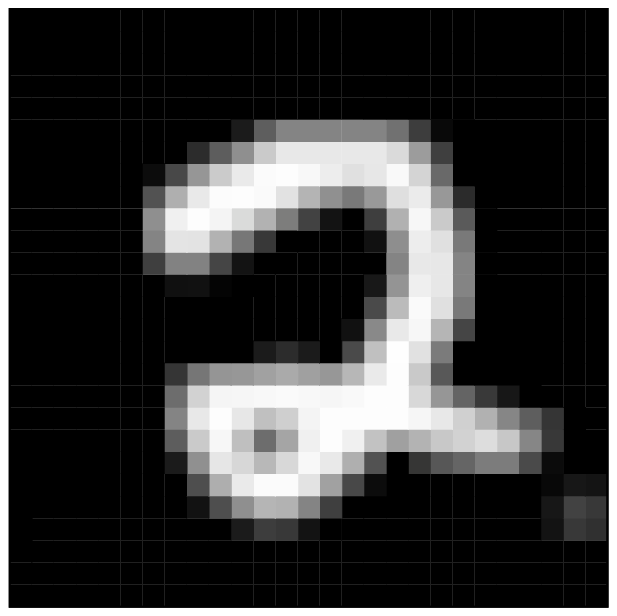}
    \includegraphics[width=.044\textwidth]{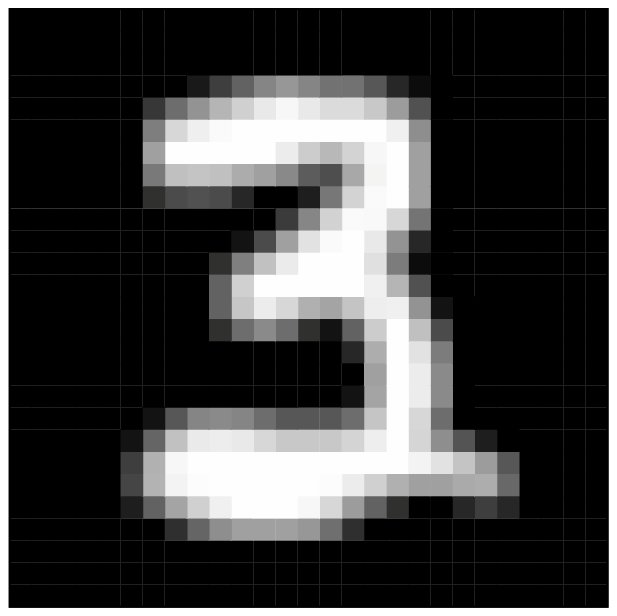}
    \includegraphics[width=.044\textwidth]{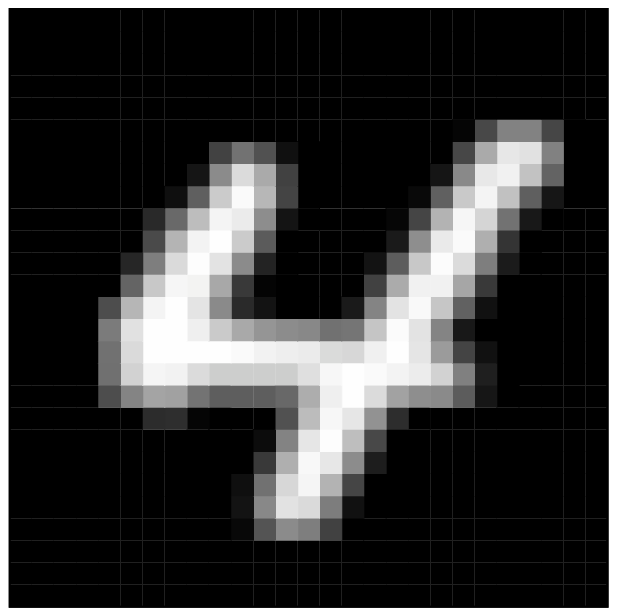}  
    \includegraphics[width=.044\textwidth]{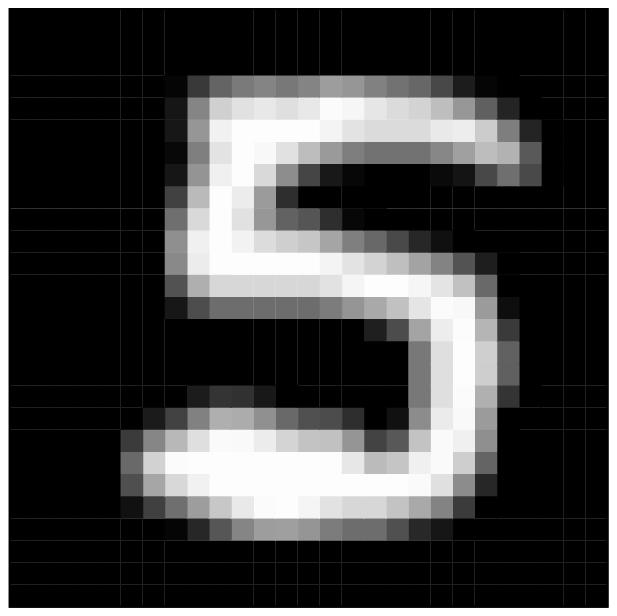}
    \includegraphics[width=.044\textwidth]{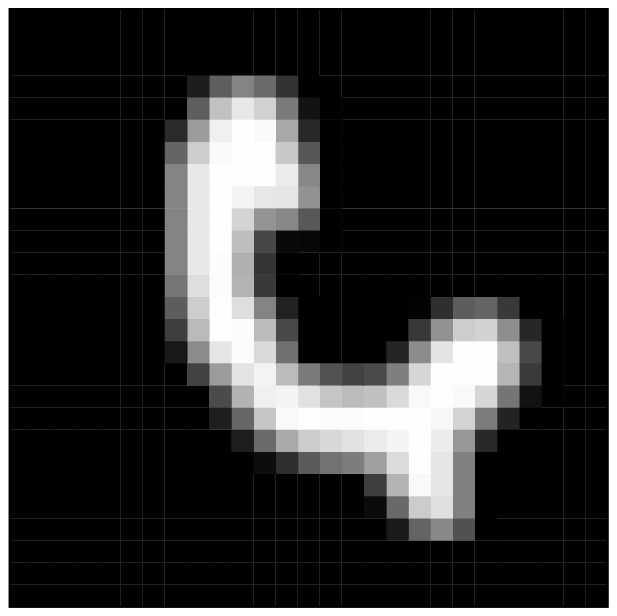}
    \includegraphics[width=.044\textwidth]{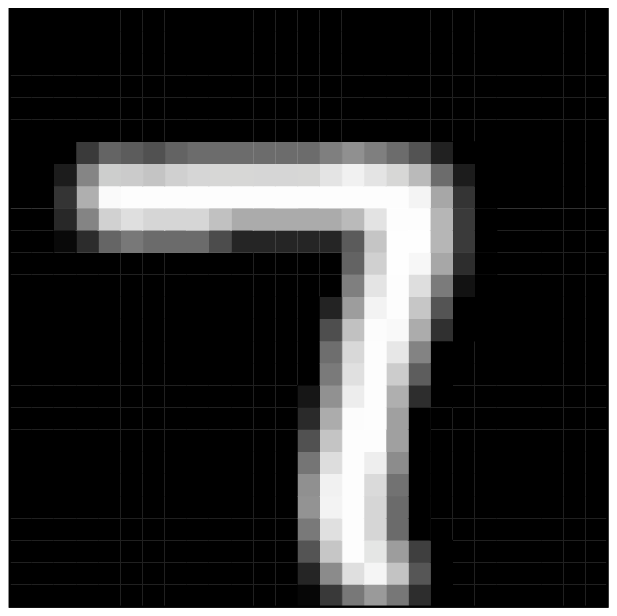}
    \includegraphics[width=.044\textwidth]{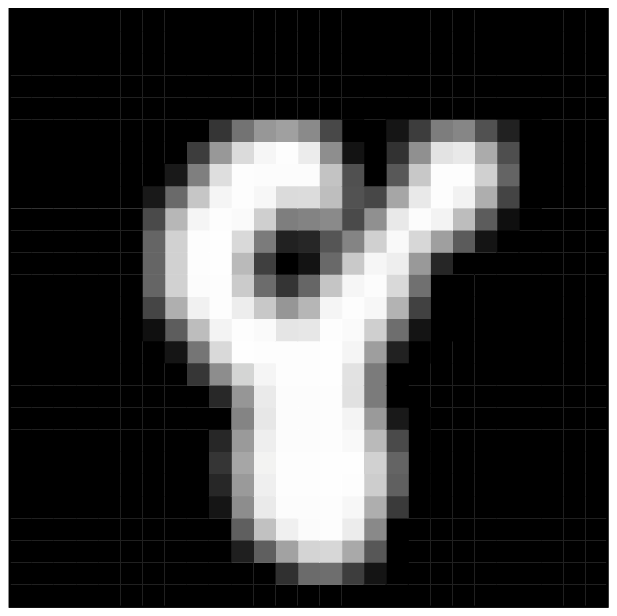}
    \includegraphics[width=.044\textwidth]{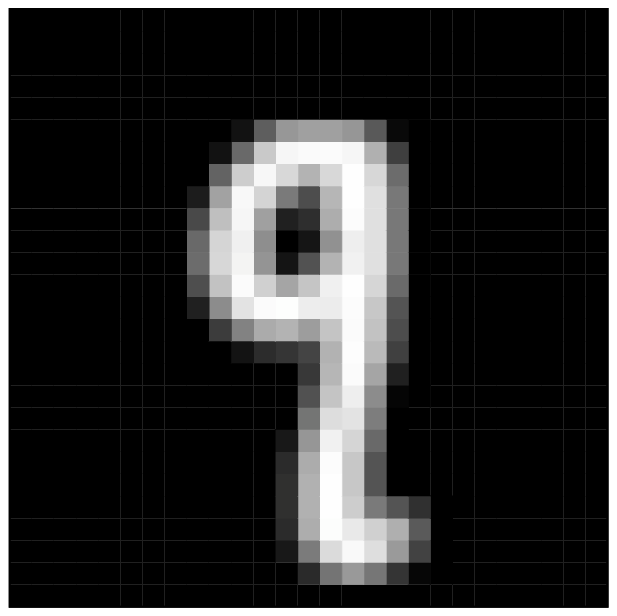}
    \label{fig:mnist}
  }
  \subfigure[Projected images in GRP-DNN]
  {
    \includegraphics[width=.044\textwidth]{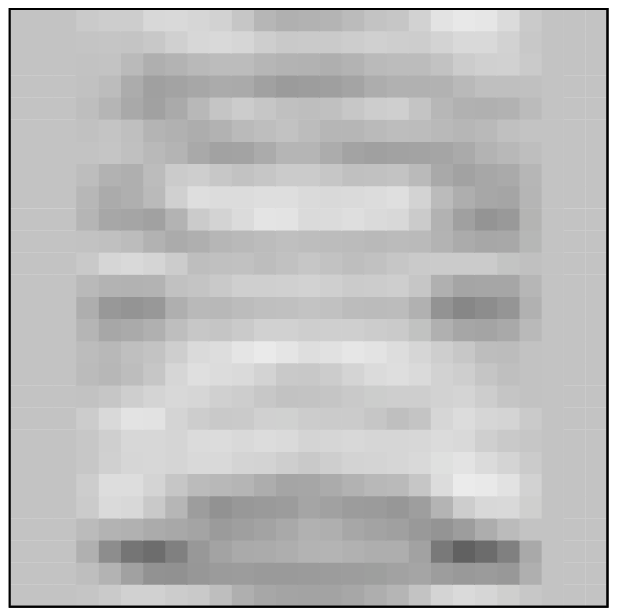}
    \includegraphics[width=.044\textwidth]{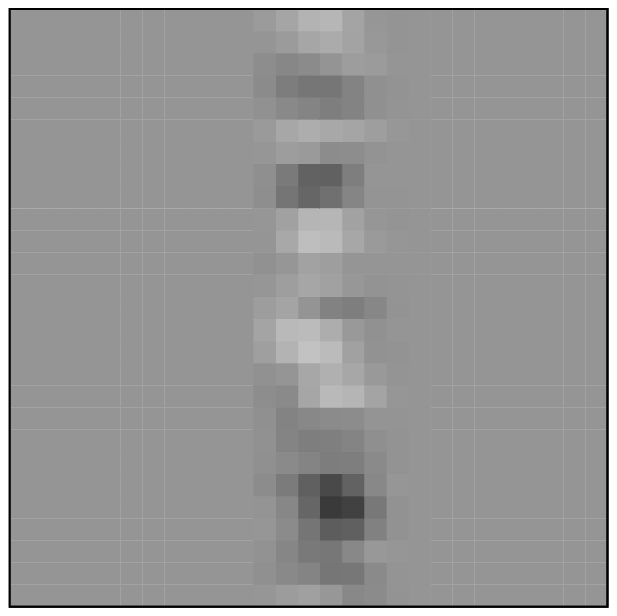}
    \includegraphics[width=.044\textwidth]{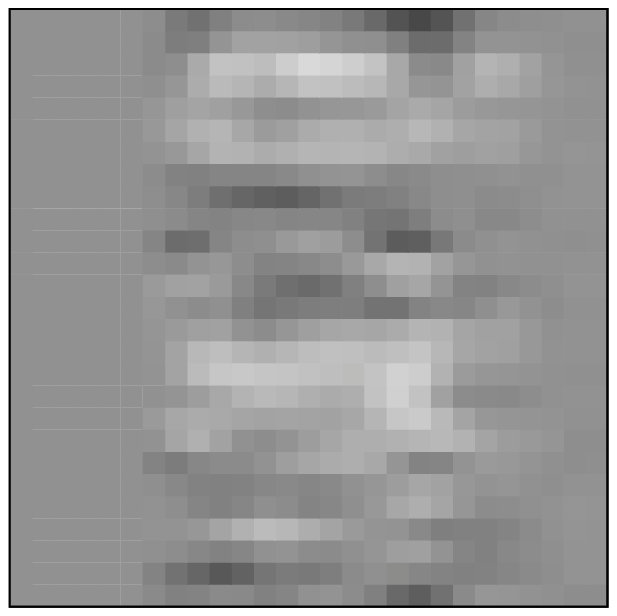}
    \includegraphics[width=.044\textwidth]{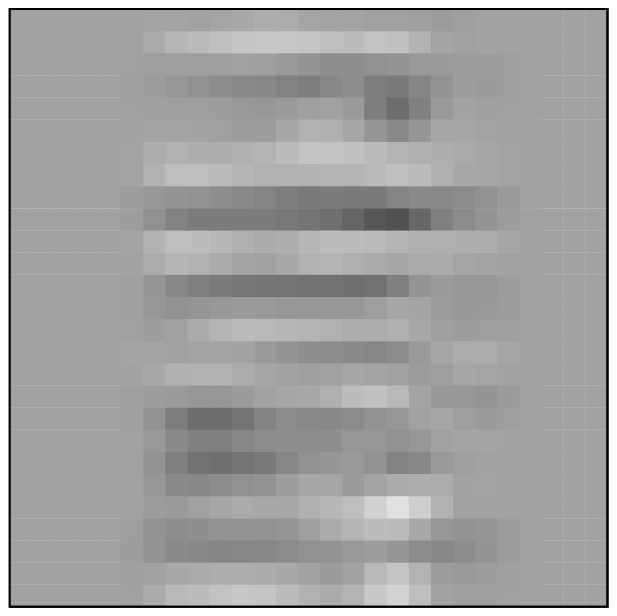}
    \includegraphics[width=.044\textwidth]{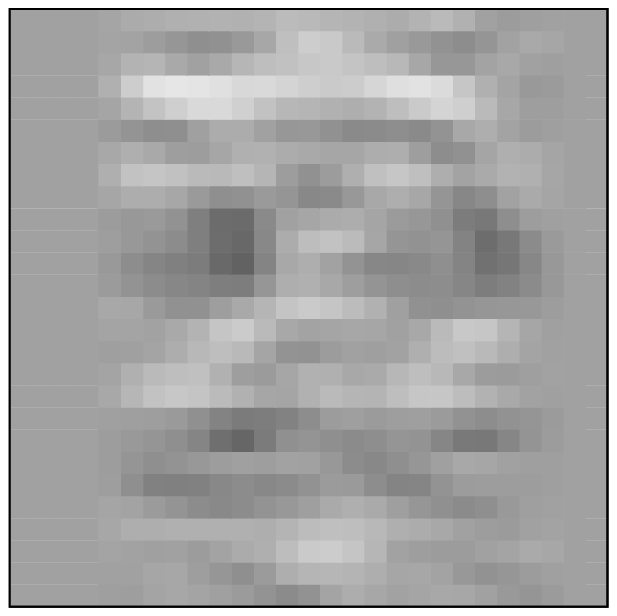}  
    \includegraphics[width=.044\textwidth]{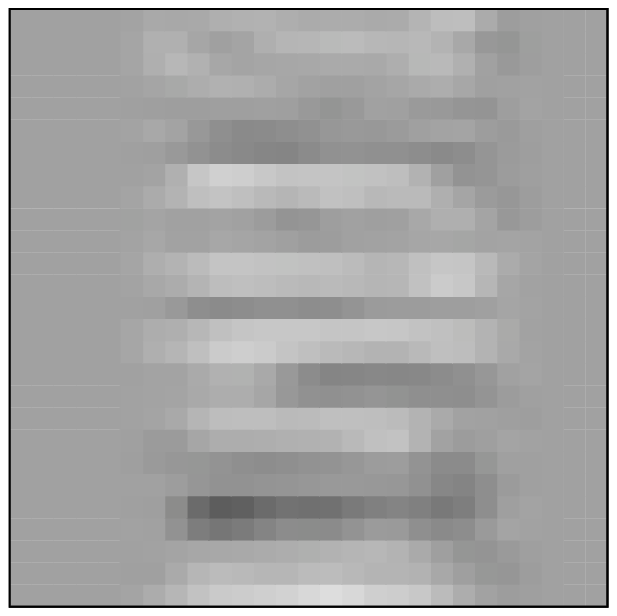}
    \includegraphics[width=.044\textwidth]{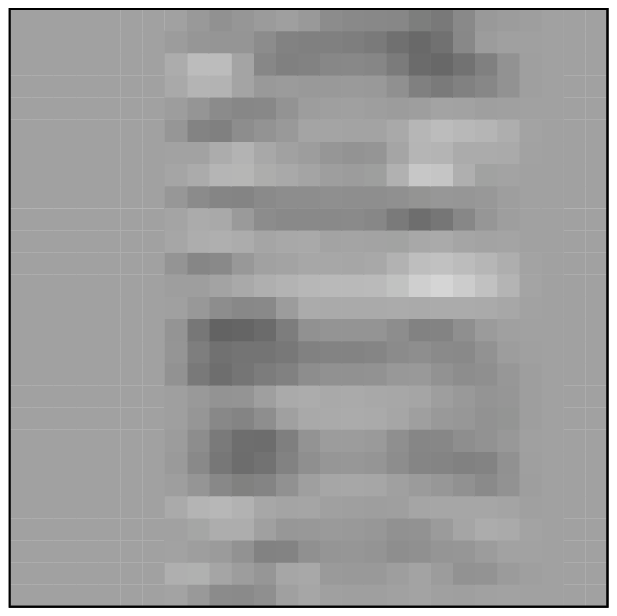}
    \includegraphics[width=.044\textwidth]{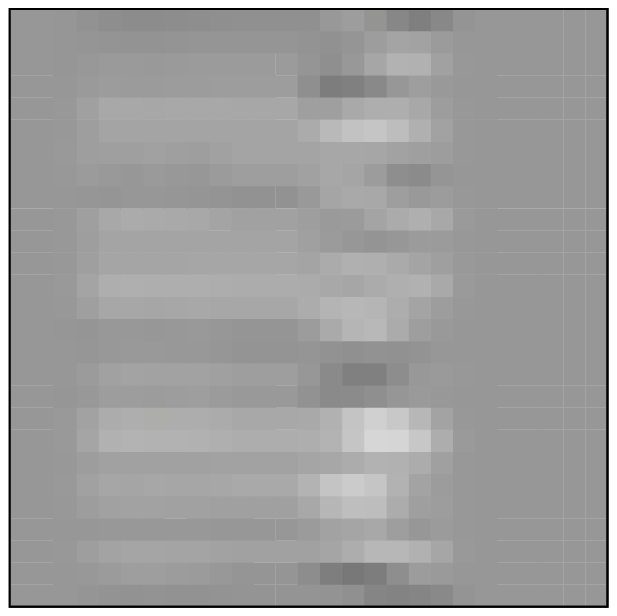}
    \includegraphics[width=.044\textwidth]{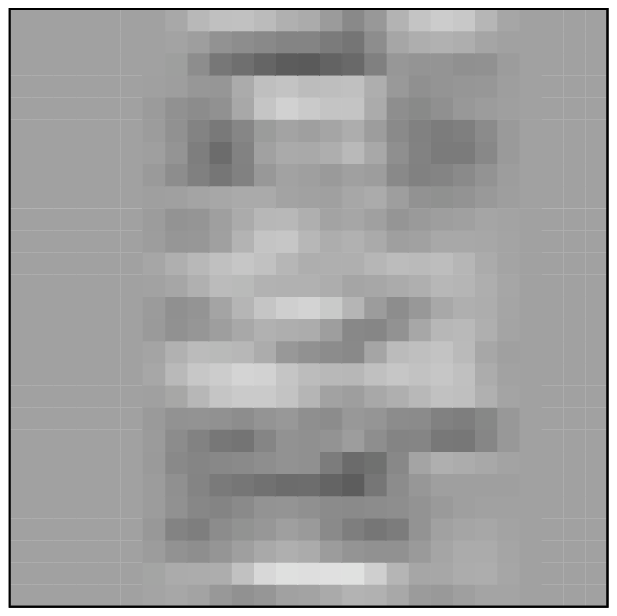}
    \includegraphics[width=.044\textwidth]{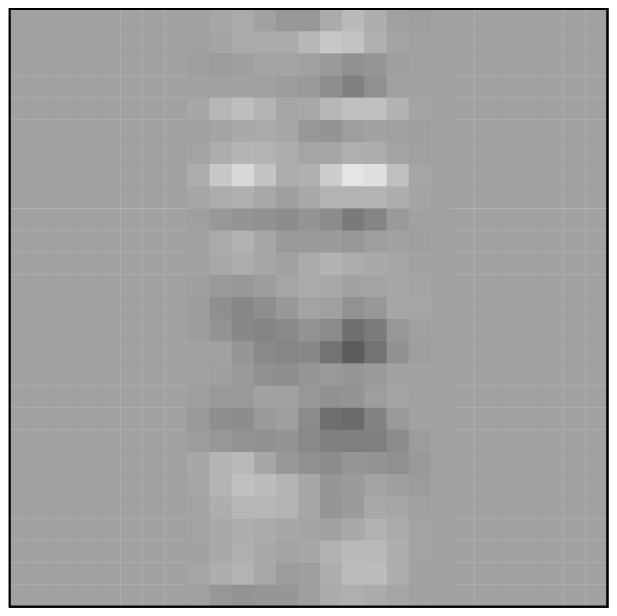}
    \label{fig:mnist-rp}
  }
  \subfigure[Noise-added images in $\epsilon$-DP-DNN ($\epsilon=50$)]
  {
    \includegraphics[width=.044\textwidth]{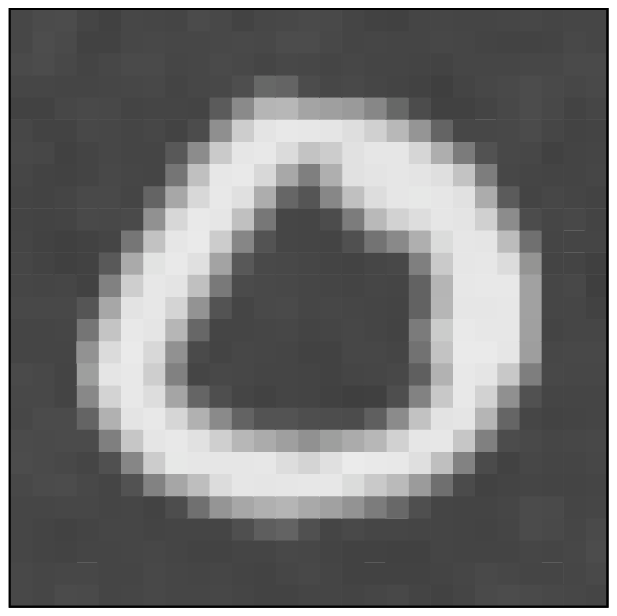}
    \includegraphics[width=.044\textwidth]{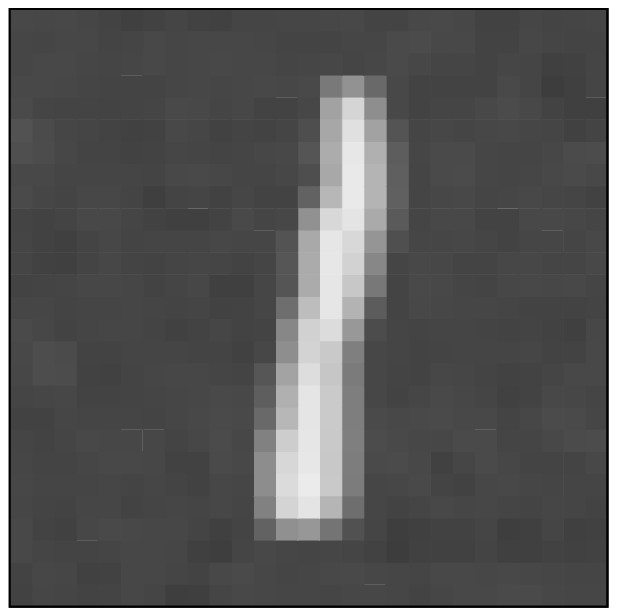}
    \includegraphics[width=.044\textwidth]{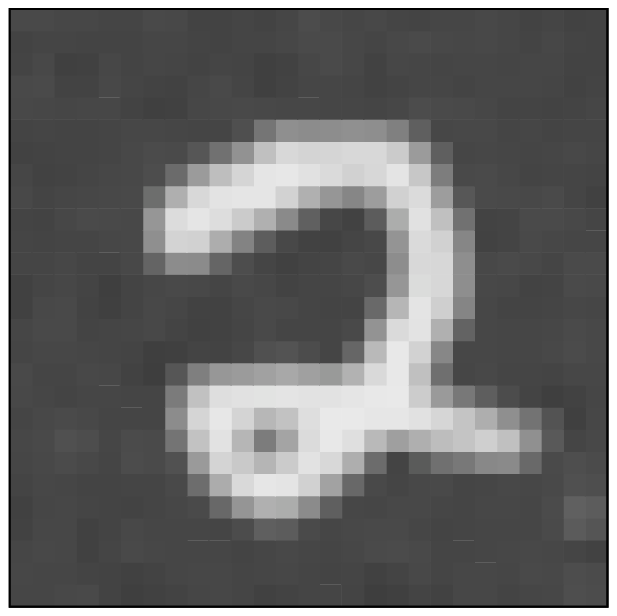}
    \includegraphics[width=.044\textwidth]{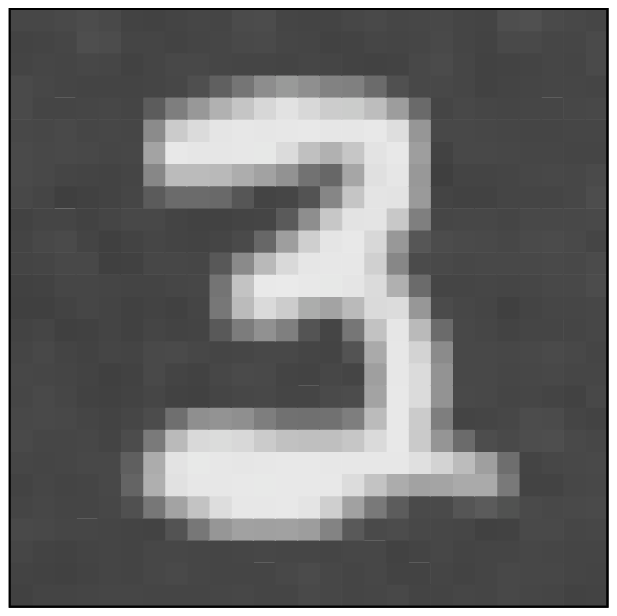}
    \includegraphics[width=.044\textwidth]{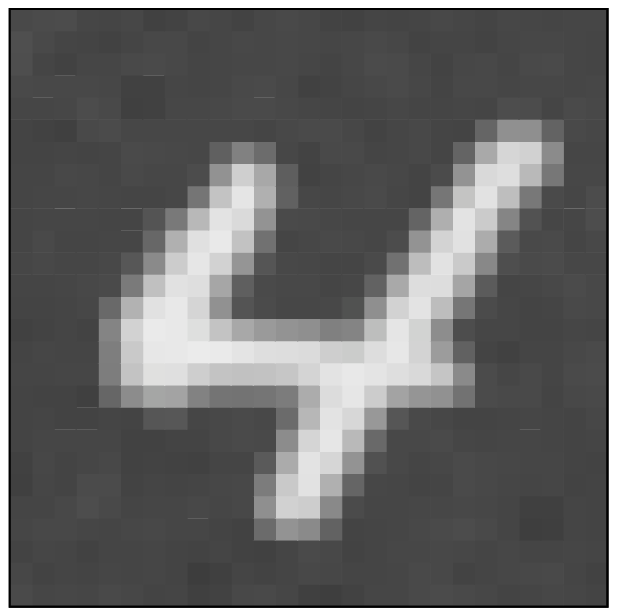}
    \includegraphics[width=.044\textwidth]{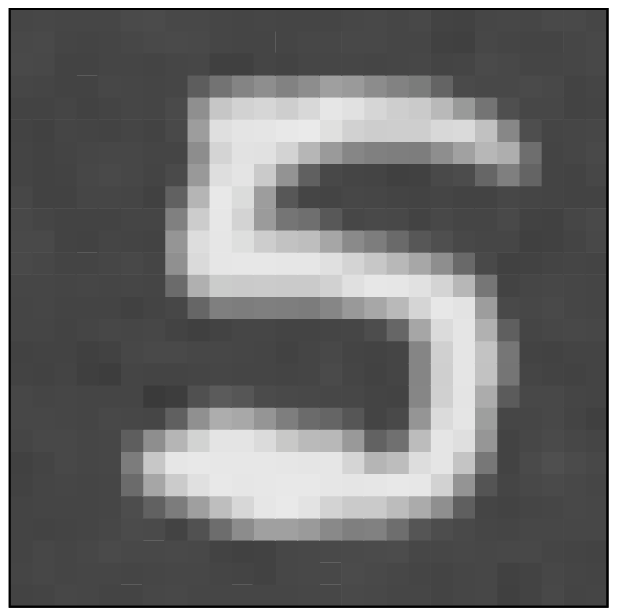}
    \includegraphics[width=.044\textwidth]{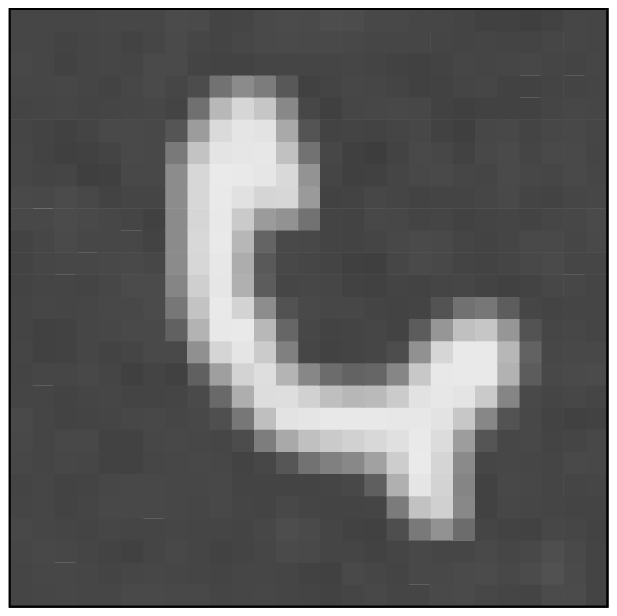}
    \includegraphics[width=.044\textwidth]{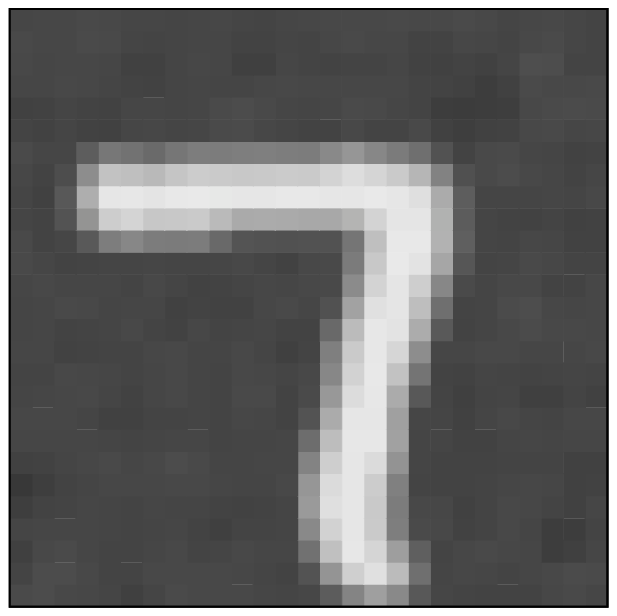}
    \includegraphics[width=.044\textwidth]{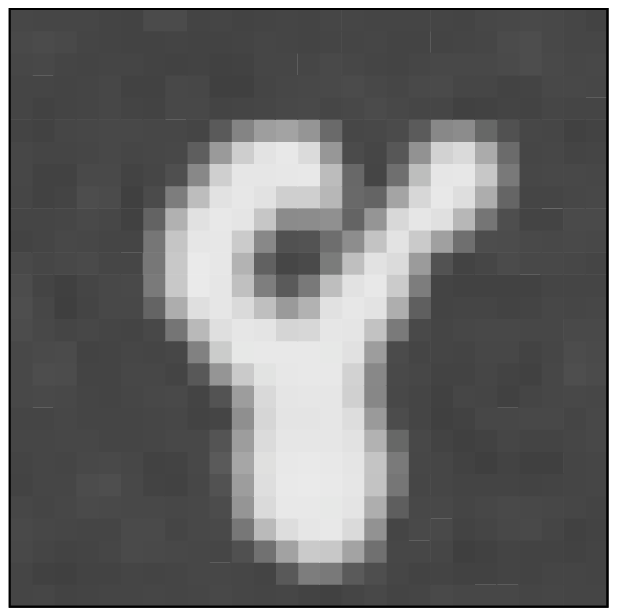}
    \includegraphics[width=.044\textwidth]{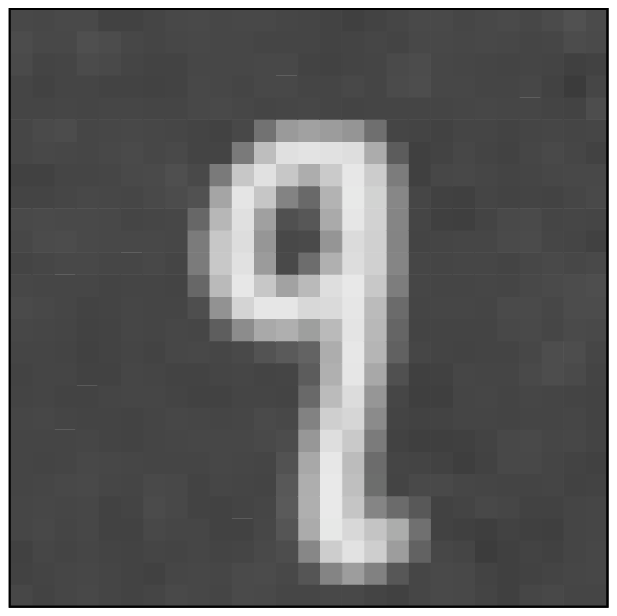}
    \label{fig:mnist-dp50}
  }
  \subfigure[Noise-added images in $\epsilon$-DP-DNN ($\epsilon=10$)]
  {
    \includegraphics[width=.044\textwidth]{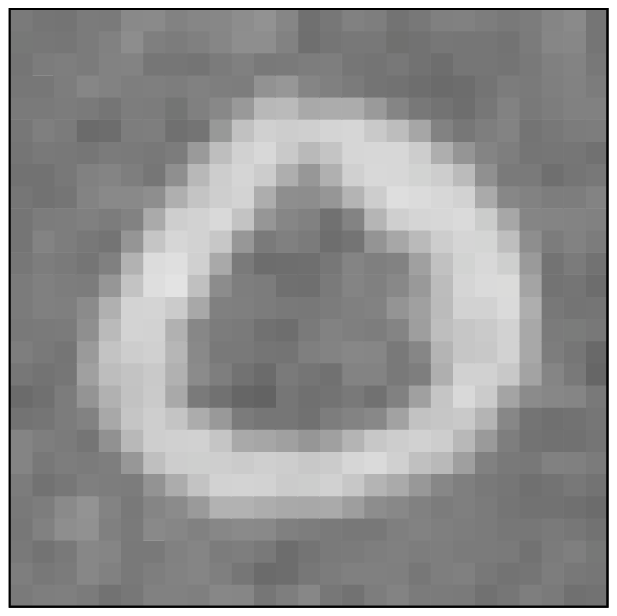}
    \includegraphics[width=.044\textwidth]{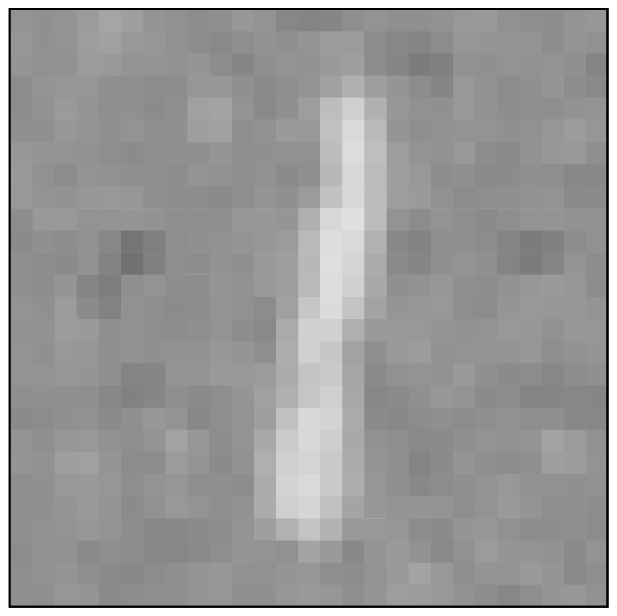}
    \includegraphics[width=.044\textwidth]{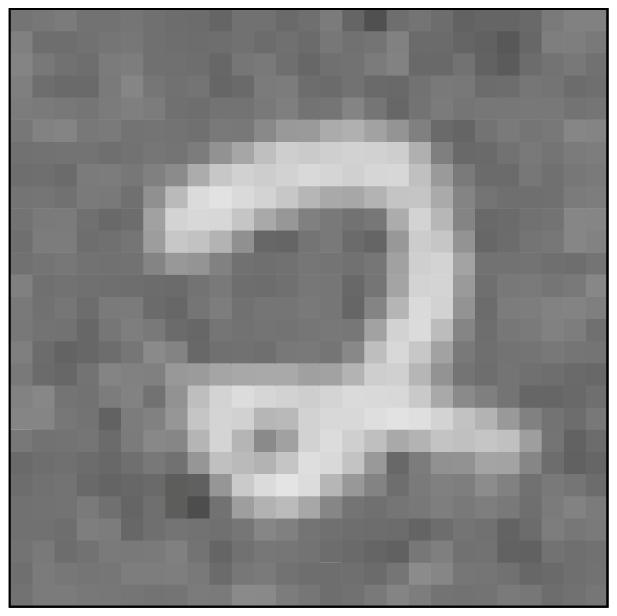}
    \includegraphics[width=.044\textwidth]{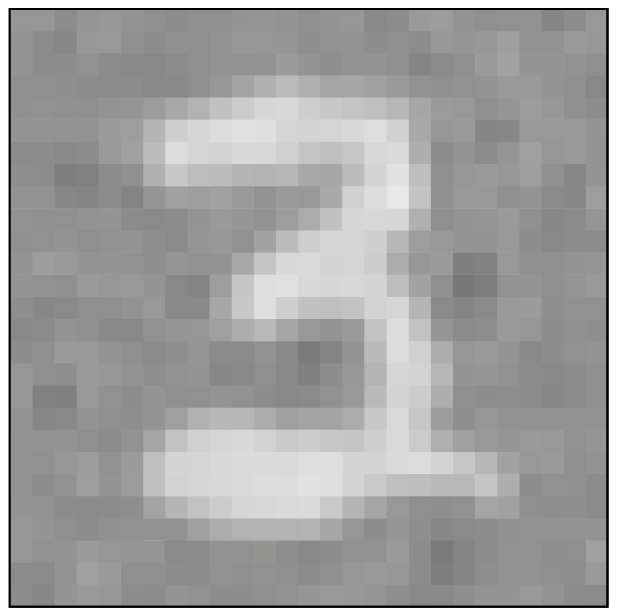}
    \includegraphics[width=.044\textwidth]{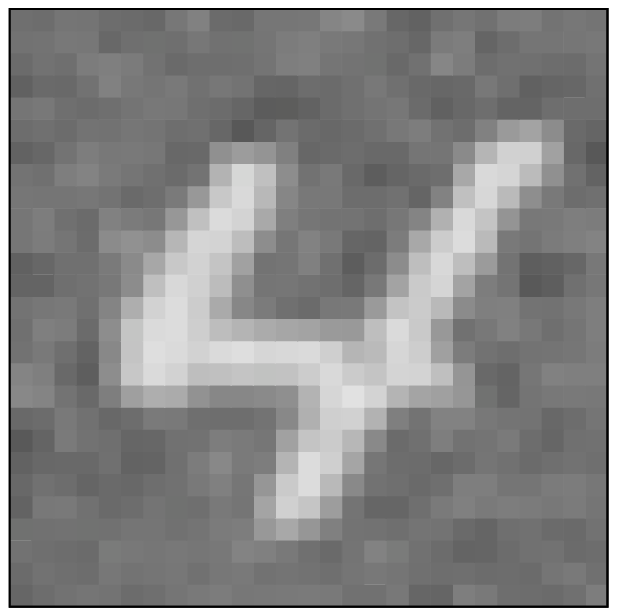}  
    \includegraphics[width=.044\textwidth]{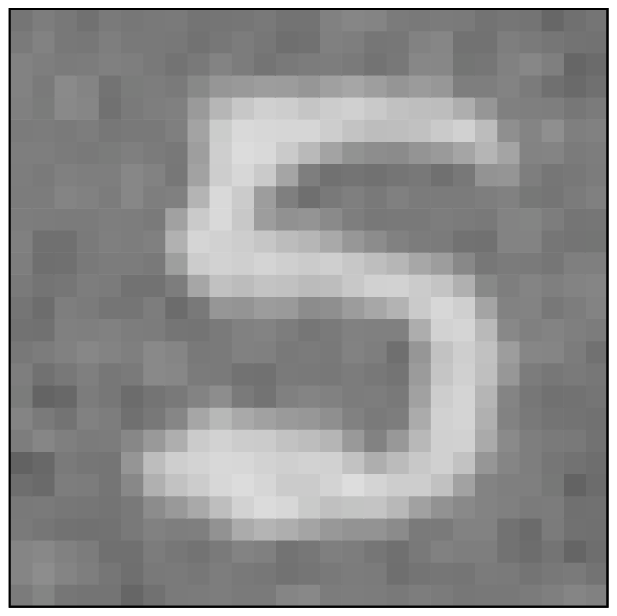}
    \includegraphics[width=.044\textwidth]{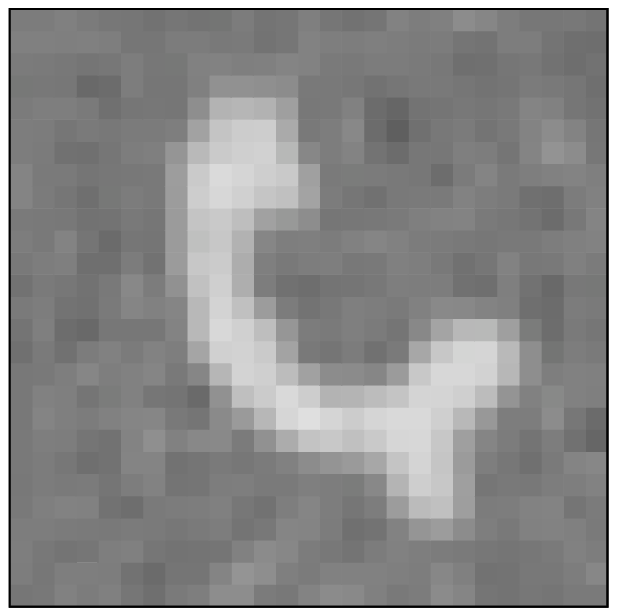}
    \includegraphics[width=.044\textwidth]{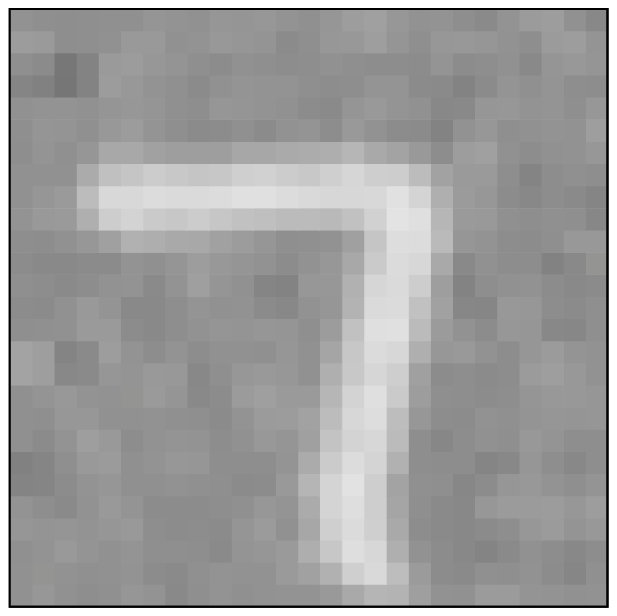}
    \includegraphics[width=.044\textwidth]{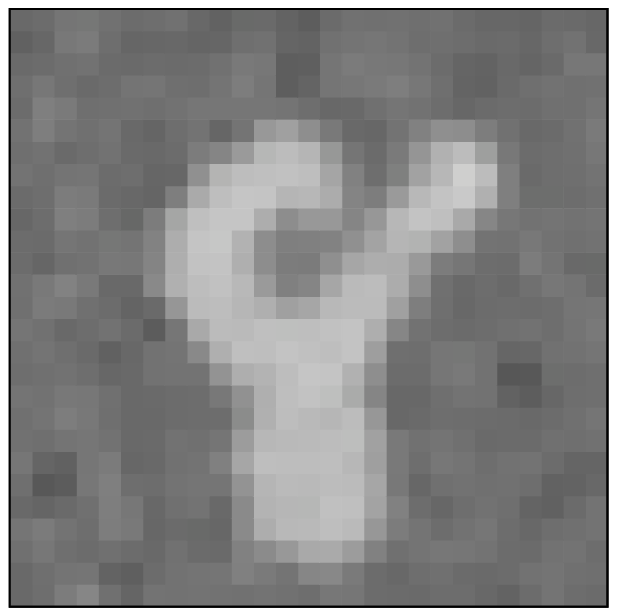}
    \includegraphics[width=.044\textwidth]{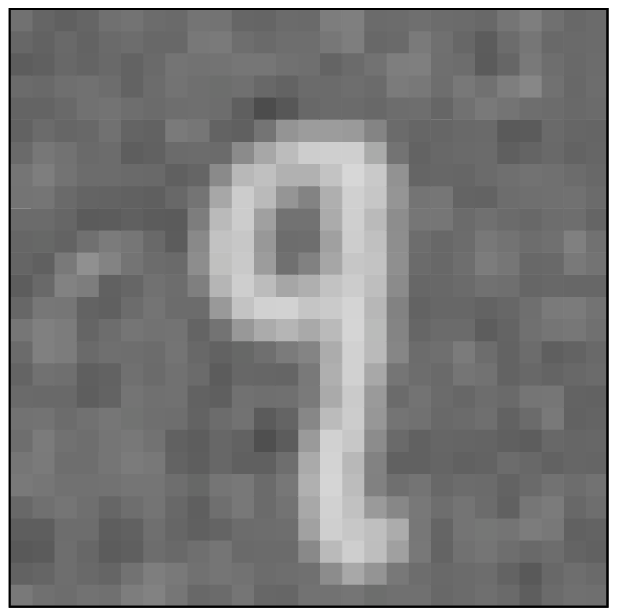}
    \label{fig:mnist-dp10}
  }
  \caption{Example images from MNIST dataset.}
  \label{fig:mnist-examples}
\end{figure}

The performance evaluation is performed based on two datasets, i.e., MNIST \cite{mnist} and spambase \cite{spambase}.
The MNIST dataset consists of 60,000 training samples and 10,000 testing samples. Each sample is a $28 \times 28$ grayscale image showing a handwritten digits from 0 to 9. Fig.~\ref{fig:mnist} shows an instance of each digit. The spambase dataset consists of 4,601 samples. Each sample consists of (i) a 57-dimensional feature vector that is extracted from an e-mail message and (ii) a class label indicating whether the e-mail message is an unsolicited commercial e-mail. The details of the feature vector can be found in \cite{spambase}. As the data volume of this spambase dataset is limited, we apply data augmentation to the spambase by adding zero-mean Gaussian noises, resulting in 40,000 training samples and 400 testing samples.
We choose these two datasets because the small sizes of the data vectors are commensurate with the limited computing and transmission capabilities of IoT end devices.

{\blue Training a spam detector based on user-contributed samples (e.g., e-mails) may cause privacy concerns. Thus, our proposed approach well fits in this case.} The choice of the vision-based character recognition task
with the MNIST dataset
allows us to leverage on the learning capabilities of the
deep models that are often designed for image classification. Moreover, by using images as the data vectors, the effect of the distortion caused by noise adding or random projection can be visualized for intuitive understanding.
{\blue Although the character
  recognition task is not privacy-sensitive, its results
  will provide understanding on other image classification-based privacy-sensitive applications, such as collaboratively training a mood classifier using the photos in the album of the users' smartphones.}


For a PPCL system with $N$ participants, we divide both the training and testing samples into $N$ disjoint sets evenly. Each set is assigned to a participant. Under GRP-DNN, GRP-SVM, and GRP-NCL, each participant independently generates its random Gaussian matrix as described in \sect\ref{subsubsec:projection} and uses the matrix to project its plaintext data vectors. The deep models and SVM are trained by the coordinator based on the projected or noise-added training data vectors from the participants. The trained deep models and SVM are used to classify the projected or noise-added testing data vectors to measure the test accuracy as the evaluation results.



\subsection{Evaluation Results with MNIST Dataset}
\label{subsec:MNIST}



\begin{figure}
\begin{tabular}{cc|c|c|c|c|c|c|c|c|c|c}
\cline{3-6} \cline{8-11}
\begin{minipage}{0.05\textwidth}
\centering
\includegraphics[width=\textwidth]{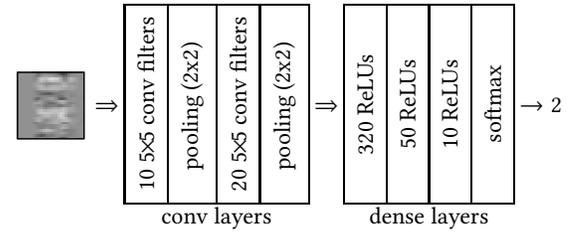}
\end{minipage}
& $\!\!\!\!\Rightarrow \!\!$
& \rotatebox[origin=c]{90}{$\;$ 10 $5\!\!\times \!\!5$ conv filters $\;$} & \rotatebox[origin=c]{90}{pooling (2x2)} & \rotatebox[origin=c]{90}{20 $5\!\! \times \!\! 5$ conv filters} & \rotatebox[origin=c]{90}{pooling (2x2)} & $\!\!\Rightarrow \!\!$ & \rotatebox[origin=c]{90}{320 ReLUs} & \rotatebox[origin=c]{90}{50 ReLUs} & \rotatebox[origin=c]{90}{10 ReLUs} & \rotatebox[origin=c]{90}{softmax} & $\!\!\rightarrow 2$\\
\cline{3-6} \cline{8-11}
\multicolumn{2}{c}{} & \multicolumn{4}{c}{conv layers} & \multicolumn{1}{c}{} & \multicolumn{4}{c}{dense layers} & \multicolumn{1}{c}{} \\
\end{tabular}
\caption{CNN with a projected MNIST image as input.}
\label{fig:CNN-MNIST}
\end{figure}

We design a CNN that is used in the GRP-DNN, GRP-NCL, and $\epsilon$-DP-DNN approaches. The CNN consists of two convolutional layers and three dense layers of ReLUs.
We apply max pooling after each convolutional layer to reduce the dimension of data after convolution. The max pooling controls overfitting effectively and improves the CNN's robustness to small spatial distortions in the input image. The last dense layer has ten ReLUs corresponding to the ten classes of MNIST.
A softmax function is used to make the classification decision based on the outputs of the last dense layer.
Fig.~\ref{fig:CNN-MNIST} illustrates the design of the CNN.
Note that, without random projection, the CNN and the SVM with grid search for kernel parameters can achieve test accuracy of 98.7\% and 98.52\%. This shows that the CNN and SVM can well capture the patterns of MNIST.

\begin{figure}
  \centering
  \includegraphics{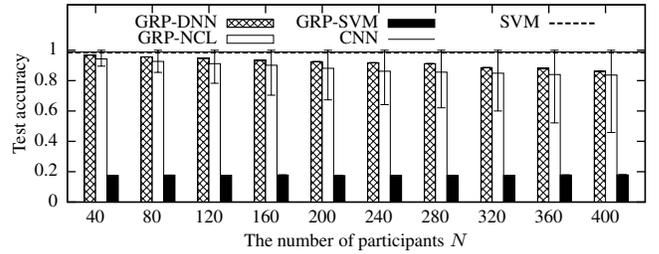}
  \caption{Impact of the number of participants (MNIST). The error bars for GRP-NCL represent min and max.}
  \label{fig:mnist-N}
\end{figure}

First, we evaluate the impact of the number of participants $N$ on the learning performance of GRP-DNN, GRP-NCL, and GRP-SVM. Fig.~\ref{fig:mnist-N} shows the results. {\blue The two horizontal lines in Fig.~\ref{fig:mnist-N} represent the test accuracy of the plain CNN and SVM without any privacy protection. The two lines overlap.} When $N$ increases from 40 to 400, the test accuracy of GRP-DNN decreases from 96.87\% to 86.18\%. If $N$ is no greater than 280, GRP-DNN can maintain a test accuracy greater than 90\%. The drop of accuracy with increased $N$ is consistent with the understanding that distinct random projection matrices increase the pattern complexity of the aggregated data. However, for MNIST data with light pattern complexities, the GRP-DNN approach can support up to 280 IoT objects for a satisfactory classification accuracy of 90\%.
Under the GRP-NCL approach, the deep models corresponding to the participants have different test accuracy values. The histogram and error bars in Fig.~\ref{fig:mnist-N} represent the average, minimum, and maximum of the test accuracy values across all trained deep models. Under each setting of $N$, the maximum test accuracy is 100\%. However, the average test accuracy is consistently lower than that of GRP-DNN. This shows that, the GRP-NCL that needs to compromise data anonymity yields inferior average learning performance compared with GRP-DNN. This result shows the advantage of collaborative learning. Lastly, the GRP-SVM approach gives poor test accuracy around 17.5\%. This is because no efficient RBF kernels can be found to create proper hyperplanes for classification. This suggests that DNNs are more efficient to cope with the distortions caused by projections.

\begin{figure}
  \centering
  \begin{minipage}[t]{.225\textwidth}
    \centering
    \includegraphics{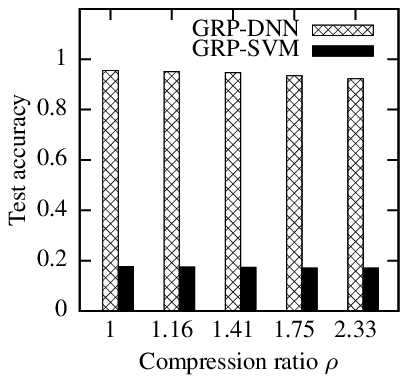}
    \caption{Impact of data compression on learning performance (MNIST, $N=100$).}
    \label{fig:mnistdata1}
  \end{minipage}
  \hspace{0.01\textwidth}
  \begin{minipage}[t]{.225\textwidth}
    \centering
    \includegraphics{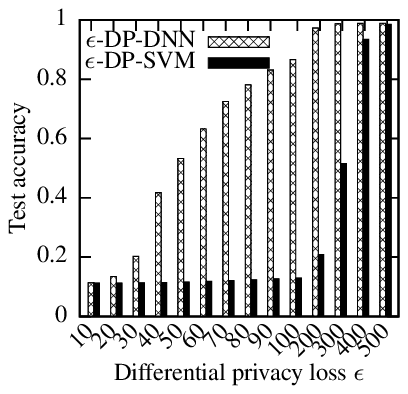}
    \caption{Impact of differential privacy loss on learning performance (MNIST).}
    \label{fig:mnist-dp-acc}
  \end{minipage}
\end{figure}

Second, we evaluate the impact of GRP's data compression on the learning performance. Fig.~\ref{fig:mnistdata1} shows the results when $N=100$. When the compression ratio increases from 1 (i.e., no compression) to 2.33 (i.e., 43\% of data volume is retained), the test accuracy of GRP-DNN decreases from 95.52\% to 92.85\% only. From our discussion in \sect\ref{subsubsec:projection}, the good tolerance of GRP-DNN against data compression is due to the high sparsity of the MNIST images.
In contrast, the GRP-SVM approach performs poorly under all compression ratio settings.





Then, we evaluate the impact of adding Laplacian noises to implement $\epsilon$-DP on the learning performance. Fig.~\ref{fig:mnist-dp-acc} shows the test accuracy of $\epsilon$-DP-DNN versus the privacy loss level $\epsilon$. When $\epsilon=100$ (small Laplacian noises and large differential privacy loss), the $\epsilon$-DP-DNN achieves a test accuracy of 86.6\%, lower than those achieved by GRP-DNN when $N$ is up to 400. When $\epsilon=10$, the performance of $\epsilon$-DP-DNN drops to 11.4\%, close to the performance of random guessing. For comparison, we visualize the projected and noise-added images with two $\epsilon$ settings in Fig.~\ref{fig:mnist-examples}. From Fig.~\ref{fig:mnist-rp}, we cannot visually interpret the projected images. However, from Figs.~\ref{fig:mnist-dp50} and \ref{fig:mnist-dp10}, the noise-added images are easily interpreted when $\epsilon$ is down to 10. Note that in our evaluation, we use the same CNN model as shown in Fig.~\ref{fig:CNN-MNIST} for the GRP-DNN, GRP-NCL, and $\epsilon$-DP-DNN approaches. We do not spend special efforts to improve the CNN design in favor of any approach; we only make sure the CNN fed with the original MNIST images achieves satisfactory performance. The poor performance of $\epsilon$-DP-DNN is consistent with the understanding that the performance of deep learning can be susceptible to small perturbations to the data vectors \cite{zheng2016improving}. There are also systematic approaches to generating adversary examples with small differences from the training samples to yield wrong classification results \cite{goodfellow15,bose2018adversarial}. Special cares are needed in the deep model design to improve robustness against human-indiscernible perturbations \cite{zheng2016improving}. Significant noises, which are required to achieve good DP protection, are still open challenges to deep learning. Thus, under the $\epsilon$-DP framework, it is challenging to achieve a desirable trade-off between the privacy protection strength and learning performance.

We discussed in \sect\ref{subsubsec:dp} that the additive noisification for $\epsilon$-DP is ineffective in achieving a good trade-off between learning performance and protecting the confidentiality of the raw forms of the training data. Now, we compare the results of GRP-DNN ($N=1$, $k = d-1$) and $\epsilon$-DP-DNN. We consider the worst case for GRP-DNN, i.e., the projection matrix $\mat{R}$ is revealed to the curious coordinator. From Property~\ref{property:2} in \sect\ref{subsec:random-projection}, the minimum norm estimate of the original data vector by the coordinator will have a per-element variance of about 410 for any MNIST image. Under this setting, GRP-DNN can achieve a test accuracy of 94.82\%. To achieve the same per-element variance of 410, the $\epsilon$ value adopted by the $\epsilon$-DP-DNN should be 18.89. Under this $\epsilon$ setting, the test accuracy of $\epsilon$-DP-DNN is 12.86\% only.

Fig.~\ref{fig:mnist-dp-acc} also shows the test accuracy of the $\epsilon$-DP-SVM approach. It performs poorly when $\epsilon \le 100$. Only when the added noises are very small under the settings of $\epsilon=400$ and $\epsilon=500$, this approach can achieve good test accuracy.


\subsection{Evaluation Results with Spambase Dataset}
\label{subsec:spambase}

\begin{figure}
  \includegraphics{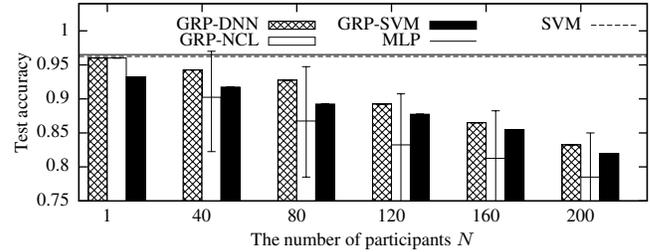}
  \caption{Impact of the number of participants (spambase). The error bars for GRP-NCL represent min and max.}
  \label{fig:spam-n}
\end{figure}

We design a 5-layer MLP classifier to detect spams. The numbers of ReLUs in the five layers are 57, 100, 50, 10, and 2, respectively. A softmax function is used lastly to make the final detection decision. Dropout is used during training to suppress overfitting. Without random projection, the MLP and the SVM with grid research for kernel parameters can achieve test accuracy of 96.52\% and 96.25\%, respectively. This shows that the MLP and SVM can well capture the patterns of spambase.

We evaluate the impact of the number of participants $N$ on the learning performance of GRP-DNN, GRP-NCL, and GRP-SVM. Fig.~\ref{fig:spam-n} shows the results. {\blue The two horizontal lines in Fig.~\ref{fig:spam-n} represent the test accuracy of the plain MLP and SVM without any privacy protection.} When $N$ increases from 1 to 200, the test accuracy of GRP-DNN decreases from 96\% to 83.25\%. If $N$ is no greater 100, GRP-DNN can maintain a test accuracy of about 90\%. The average test accuracy of GRP-NCL is about 5\% lower than that of the GRP-DNN, because GRP-NCL misses the advantages of collaborative learning. The test accuracy of the GRP-SVM is about 1.25\% to 2.75\% lower than that of the GRP-DNN. Thus, the GRP-SVM performs satisfactorily for this spambase dataset. The reasons are two-fold. First, in this spambase dataset, the classifiers operate on the e-mail features, rather than the raw data. Second, the RBF kernel is effective in capturing the features. In fact, the nature of this spambase dataset is similar to that of the 2-dimensional and 10-dimensional generated feature datasets used in \sect\ref{subsec:example}, on which the GRP-DNN and GRP-SVM perform similarly.

\subsection{Summary and Discussion}

We have several observations from the results in \sect\ref{subsec:MNIST} and \sect\ref{subsec:spambase}:
\begin{itemize}
\item Compared with SVM, deep learning can better adapt to the complexity introduced by the multiplicative projections.
\item Although the GRP-NCL approach additionally uses the identities of the participants, it gives inferior performance compared with the collaborative GRP-DNN. This shows the advantage of collaborative learning even with the privacy preservation requirement.
\item Compared with GRP-DNN, the additive noisification for $\epsilon$-DP achieves inferior trade-off between learning performance and protecting confidentiality of raw forms of training data.
\item GRP-DNN shows promising scalability with the number of participants observing low-complexity data patterns. For the MNIST and spambase datasets, the GRP-DNN can well support 100 participants with a few percents test accuracy drop. For large-scale PPCL systems involving more participants, we envision a two-tier system architecture as follows. The participants are divided into groups. At the first tier, our GRP-DNN is applied within each group; at the second tier, the DML approach is applied among the group coordinators.
\end{itemize}



%% file: implementation.tex
\section{implementation and Benchmark}
\label{sec:implementation}

In this section, we measure the overhead of two PPCL approaches (i.e., our GRP-DNN and Crowd-ML \cite{Hamm15}) and a privacy-preserving classification outsourcing approach (i.e., CryptoNets \cite{gilad2016cryptonets}) on a testbed of 14 Raspberry Pi 2 Model B nodes \cite{pi} and a powerful workstation computer. The Raspberry Pi nodes act as PPCL participants and the workstation acts as the coordinator. They are interconnected using a 24-port network switch. We benchmark these approaches using the MNIST dataset. The training and testing samples are evenly allocated to the participants, resulting in 4,285 training samples and 714 testing samples on each participant. {\blue The implementations of the three approaches (GRP-DNN, Crowd-ML, CryptoNets) on the same platform, i.e., Raspberry Pi, allow fair comparisons. The participant part of our GRP-DNN can be implemented on mote-class platforms. Our previous work \cite{tan2017joint} has implemented Gaussian matrix generation and GRP on the MSP430-based Kmote platform. However, it is difficult/impossible to implement Crowd-ML and CryptoNets on mote-class platforms.}


We implement our GRP-DNN approach on the testbed. The compression ratio $\rho=1$ (i.e., no compression). Table~\ref{tab:performance} shows the benchmark results. During the training phase, each GRP-DNN participant needs to transmit a total of $33.6\,\text{MB}$ projected data.
A participant can complete projecting all the 4,285 training images within $0.96\,\text{s}$. The coordinator needs $928.34\,\text{s}$ to train the CNN. In our GRP-DNN implementation, the testing phase is performed on the coordinator. During the testing phase, each participant completes projecting all the 714 testing images within $0.16\,\text{s}$ and transmits a total of $5.6\,\text{MB}$ data to the coordinator. The coordinator needs $40.88\,\text{s}$ to classify all projected testing images from the participants. Note that GPU acceleration is not used in this benchmark for GRP-DNN during both the training and testing phases.

\begin{table}
  \small
  \caption{The overhead of various approaches.}
  \label{tab:performance}
  \begin{tabular}{c|c|c|c|c}
    \hline
    & Overhead & GRP-DNN & Crowd-ML & $\!\!$CryptoNets$\!\!$ \\
    \hline
    \multirow{3}{*}{\rotatebox[origin=c]{90}{Training}} & Participant comm. vol. & $33.6\,\text{MB}$ & $117.2\,\text{MB}$ & n/a \\
    & Participant compute time & $0.96\,\text{s}$ & $367.24\,\text{s}$  & n/a \\
    & $\!\!$Coordinator compute time$\!\!$ & $928.34\,\text{s}$ & $1.04\,\text{s}$ & n/a \\
    \hline
    \multirow{3}{*}{\rotatebox[origin=c]{90}{Testing}} & Participant comm. vol. & $5.6\,\text{MB}$ & n/a & $15.0\,\text{MB}$ \\
    & Participant compute time & $0.16\,\text{s}$ & $4.67\,\text{s}$ & 116 hours \\
    & $\!\!$Coordinator compute time$\!\!$ & $40.88\,\text{s}$ & n/a & \\
    \hline
    \multicolumn{5}{l}{n/a represents ``not applicable.''}
  \end{tabular}
\end{table}


The Crowd-ML \cite{Hamm15} is a DML approach. In Crowd-ML, a participant checks out the global classifier parameters from the coordinator and computes the gradients using its own training data. Then, the participants transmit the gradients to the coordinator that will update the global classifier parameters. Thus, during the training phase, the participants and the coordinator repeatedly exchange parameters. We apply an existing implementation of Crowd-ML \cite{crowd-ml-impl} on our testbed.
Our measurement shows that, during the training phase, each participant needs to upload and download a total of $117.2\,\text{MB}$ data, which is 3.5x of our GRP-DNN. {\blue The participant compute time is more than 350x of that under GRP-DNN.} Despite the larger volume of data exchanges, Crowd-ML achieves 91.28\% test accuracy only, which is lower than the 95.58\% test accuracy achieved by GRP-DNN. This is because Crowd-ML uses a simple multiclass logistic classifier, which is inferior compared with the CNN used by GRP-DNN in terms of learning performance. Note that during the testing phase of Crowd-ML, the participants execute their local classifiers. Thus, they do not need to transmit the testing samples to the coordinator for classification.


CryptoNets \cite{gilad2016cryptonets} uses homomorphic encryption algorithm to encrypt a testing sample during the classification phase and transmits the encrypted sample to the coordinator. Then, the coordinator uses a neural network trained with plaintext data to classify the encrypted testing sample. We have implemented the homomorphic encryption part of CrytoNets that runs on the Raspberry Pis.
The volume of the 714 encrypted testing images is $15\,\text{MB}$, almost 3x of the data volume generated by random projection. {\blue In particular, a Raspberry Pi node takes about 10 minutes and a total of 116 hours to encrypt an image and all the testing images, respectively. This is 2.6 million times slower than the random projection computation.} This result clearly shows that the high computation complexity of the homomorphic encryption makes CryptoNets ill-suited for resource-constrained devices.



%% file: conclude.tex
\section{Conclusion}
\label{sec:conclude}

This paper proposes a practical privacy-preserving collaborative learning approach, in which the resource-constrained learning participants apply independent Gaussian projections on their training data vectors and the coordinator applies deep learning to train a classifier based on the projected data vectors. Our approach protects the confidentiality of the raw forms of the training data against the honest-but-curious coordinator. Evaluation using two datasets shows that our approach outperforms various baselines and exhibits promising scalability with respect to the number of participants observing low-complexity data patterns. Benchmark on a testbed shows the practicality and efficiency of our approach.
